\documentclass{article}

\usepackage{amsmath,amsfonts,bm}

\def\eqref#1{equation~\ref{#1}}

\def\1{\bm{1}}

\def\eps{{\epsilon}}

\DeclareMathAlphabet{\mathsfit}{\encodingdefault}{\sfdefault}{m}{sl}
\SetMathAlphabet{\mathsfit}{bold}{\encodingdefault}{\sfdefault}{bx}{n}

\usepackage[utf8]{inputenc} 
\usepackage[T1]{fontenc}    
\usepackage{url}            
\usepackage{booktabs}       
\usepackage{amsfonts}       
\usepackage{nicefrac}       
\usepackage{empheq}
\usepackage{natbib}
\usepackage{amsthm}
\usepackage{caption}
\usepackage{mathrsfs}
\usepackage{mathtools}
\usepackage{amssymb}
\usepackage{tikz}
\usepackage{multirow}
\usepackage{subcaption}
\usepackage{tikz}
\usepackage{pgfplots}
\usepgfplotslibrary{fillbetween}
\tikzset{>=latex}
\pgfplotsset{compat=1.16}
\usepackage{tkz-euclide}
\usepackage{xcolor}
\usetikzlibrary{calligraphy, calc, fit}
\usepackage{ifthen}
\usepackage{tabularx}
\usetikzlibrary{calc,positioning,matrix, decorations.pathreplacing}

\usepackage[colorlinks = true,
            linkcolor = blue,
            urlcolor  = blue,
            citecolor = blue]{hyperref} 
\usepackage{xr}

\theoremstyle{definition}
\newtheorem{definition}{Definition}

\theoremstyle{remark}

\newcommand{\reals}{\mathbb{R}}
\newcommand{\naturals}{\mathbb{N}}
\newcommand{\bigO}{\mathcal{O}}
\newcommand{\ttfrac}[2]{\nicefrac{#1}{#2}}
\newcommand{\tseries}[1]{\mathcal{S}(#1)}

\newcommand{\brownian}[7]{
    \draw[#4,line width=0.5mm] (#6, #7)
    \foreach \x in {1,...,#1}
        {   
            -- ++(#2, rand*#3)
        }
        node[right] {#5};
}

\newcommand{\origin}[3]{ 
    \tkzDefPoint(#1, #2){B}
    \tkzLabelPoint[left](B){#3}
    \node at (B) [circle,fill,inner sep=2pt]{};
}

\def\seed{2}
\def\graphlen{8}
\def\xgap{2}
\def\curveshift{1}

\def\yd{\graphlen/2}
\def\yu{\graphlen/2}

\def\Bs{\graphlen + \xgap}
\def\Cs{2*\graphlen + 2*\xgap}
\def\Ds{3*\graphlen + 3*\xgap}
\def\Es{4*\graphlen + 4*\xgap}

\def\numpoints{50}
\def\pointdistance{\graphlen / \numpoints - 1/\numpoints}
\def\startone{\graphlen * 0.2}
\def\starttwo{\graphlen * 0}
\def\startthree{-\graphlen * 0.1}
\def\randone{0.6}
\def\randtwo{0.5}
\def\randthree{0.6}

\colorlet{lightgray}{gray!10}
\definecolor{colorp1}{HTML}{003f5c}
\definecolor{colorp2}{HTML}{58508d}
\definecolor{colorp3}{HTML}{ff6361}
\definecolor{colord1}{HTML}{66C2A5}
\definecolor{colord2}{HTML}{8DA0CB}
\definecolor{colord3}{HTML}{FC8D62}

\definecolor{colorp1}{HTML}{4878d0}
\definecolor{colorp2}{HTML}{ee854a}
\definecolor{colorp3}{HTML}{6acc64}
\definecolor{colord1}{HTML}{003f5c}
\definecolor{colord2}{HTML}{bc5090}
\definecolor{colord3}{HTML}{ff6361}

\tikzset{
    ncbar angle/.initial=90,
    ncbar/.style={
        to path=(\tikztostart)
        -- ($(\tikztostart)!#1!\pgfkeysvalueof{/tikz/ncbar angle}:(\tikztotarget)$)
        -- ($(\tikztotarget)!($(\tikztostart)!#1!\pgfkeysvalueof{/tikz/ncbar angle}:(\tikztotarget)$)!\pgfkeysvalueof{/tikz/ncbar angle}:(\tikztostart)$)
        -- (\tikztotarget)
    },
    ncbar/.default=0.5cm,
}

\tikzset{square left brace/.style={ncbar=0.5cm}}
\tikzset{square right brace/.style={ncbar=-0.5cm}}

\usepackage{hyperref}

\usepackage[accepted]{style/icml2021}

\icmltitlerunning{A Generalised Signature Method for Multivariate Time Series Feature Extraction}

\usepackage{style/widetext}

\begin{document}

\twocolumn[
\icmltitle{A Generalised Signature Method for Multivariate Time Series Feature Extraction}



\icmlsetsymbol{equal}{*}

\begin{icmlauthorlist}
\icmlauthor{James Morrill}{to,goo,equal}
\icmlauthor{Adeline Fermanian}{sor,equal}
\icmlauthor{Patrick Kidger}{to,goo,equal}
\icmlauthor{Terry Lyons}{to,goo}

\end{icmlauthorlist}

\icmlaffiliation{to}{Mathematical Institute, University of Oxford, UK}
\icmlaffiliation{goo}{The Alan Turing Institute, British Library, UK}
\icmlaffiliation{sor}{Sorbonne Universit{\'e}}
\icmlsetsymbol{equal}{*}

\icmlcorrespondingauthor{James Morrill}{morrill@maths.ox.ac.uk}

\icmlkeywords{Machine Learning, ICML, Signatures, Time Series}

\vskip 0.3in
]



\printAffiliationsAndNotice{\icmlEqualContribution} 

\begin{abstract}
The `signature method’ refers to a collection of feature extraction techniques for multivariate time series, derived from the theory of controlled differential equations. There is a great deal of flexibility as to how this method can be applied. On the one hand, this flexibility allows the method to be tailored to specific problems, but on the other hand, can make precise application challenging. This paper makes two contributions. First, the variations on the signature method are unified into a general approach, the \emph{generalised signature method}, of which previous variations are special cases. A primary aim of this unifying framework is to make the signature method more accessible to any machine learning practitioner, whereas it is now mostly used by specialists. Second, and within this framework, we derive a canonical collection of choices that provide a domain-agnostic starting point. We derive these choices as a result of an extensive empirical study on 26 datasets and go on to show competitive performance against current benchmarks for multivariate time series classification. Finally, to ease practical application, we make our techniques available as part of the open-source [redacted] project.
\end{abstract}

\section{Introduction}
	A multivariate time series is obtained by observing $d$ quantities evolving with time, which can be written as an array $\mathbf{x}=(x_1, \dots,x_n)$, where $n$ is the length of the series, and $x_i \in \reals^d$ for each $i \in \{1, \dots,n\}$. These data are common in various fields (finance, health, energy...) and offer several specific challenges: they are often highly dimensional, as both the number of channels $d$ and the length of the series $n$ may be large, the values $x_i$ are correlated, and the different channels may interact. Finally, the inputs may be of different length and the data may be irregularly sampled.
	
	One approach is to construct models that directly accept some of these issues; for example recurrent neural networks handle correlated inputs with varying lengths. A second option is to use feature extraction techniques, which normalise the data so that other techniques may then be applied.  Methods such as the shapelet transform \citep{ye2009firstshapelet, grabocka2014learningshapelet, kidger2020shapelets}, Gaussian process adapters \citep{gp-adapter1, gp-adapter2, pathsig}, and in particular the signature method \citep{levin2013learning}, all fit into this category. 
	
\colorlet{lightgray}{gray!10}
\definecolor{colorp1}{HTML}{4878d0}
\definecolor{colorp2}{HTML}{ee854a}
\definecolor{colorp3}{HTML}{6acc64}
\definecolor{colord1}{HTML}{003f5c}
\definecolor{colord2}{HTML}{bc5090}
\definecolor{colord3}{HTML}{ff6361}
\definecolor{colorls1}{HTML}{66C2A5}
\definecolor{colorls2}{HTML}{8DA0CB}
\definecolor{colorls3}{HTML}{FC8D62}

\global\def\plotrange{0, 1, 4, 8, 11, 15, 18, 19}
\def\myarr {
    (0.131579, 0.5)
    (0.2631579, 1.228486295)
    (1.05263158, 2.37243038 - 0.3)
    (2.10526316, 2.232905675 - 0.3)
    (2.89473684 - 0.1, 1.597253615)
    (3.94736842 - 0.3, 1.09415549 + 0.2)
    (4.7368421 - 0.5, 1.6927394599999999)
    (5.0 -0.3, 2.5)
}

\global\def\xs{{0.131579, 0.2631579, 0.52631578, 0.78947368, 1.05263158, 1.31578948, 1.57894736, 1.84210526, 2.10526316, 2.36842106, 2.63157894, 2.89473684 - 0.1, 3.15789474, 3.42105264, 3.68421052, 3.94736842 - 0.3, 4.21052632, 4.47368422, 4.7368421 - 0.5, 5.0 - 0.3}}

\global\def\ys{{0.5, 1.228486295, 1.7697186200000001, 2.14419923, 2.37243038 - 0.3, 2.47491434 - 0.3, 2.47215338, 2.384649725, 2.232905675 - 0.3, 2.037423455, 1.81870535, 1.597253615, 1.39357049, 1.2281582599999998, 1.121519165, 1.09415549 + 0.2, 1.166569475, 1.359263375, 1.6927394599999999, 2.5}}

\global\def\xlen{19}

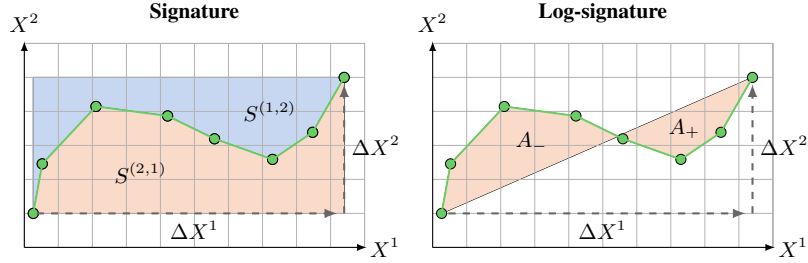
\begin{figure*}[t]
    \centering
    \resizebox{0.64\linewidth}{!}{
        \begin{tikzpicture}
            \small
            \draw[xshift=0cm, name path=one] plot coordinates {
                \myarr
            };
            \draw[xshift=0cm, name path=two] plot coordinates {
                (\xs[0], \ys[0]) (\xs[0], \ys[\xlen]) (\xs[\xlen], \ys[\xlen])
            };
            \tikzfillbetween[
                of=one and two, split
            ] {pattern=north west lines, colorp1!30};
            \draw[xshift=0cm, name path=one] plot coordinates {
                \myarr
            };
            \draw[xshift=0cm, name path=two] plot coordinates {
                (\xs[0], \ys[0]) (\xs[\xlen], \ys[0]) (\xs[\xlen], \ys[\xlen])
            };
            \tikzfillbetween[
                of=one and two, split
            ] {pattern=north west lines, colorp2!30};
            \draw[step=0.5, draw=black!30] (0, 0) grid (5, 3) rectangle (0, 0);
            \foreach \i in \plotrange {
                \node at (\xs[\i], \ys[\i])[circle, draw=black, fill=colorp3, inner sep=1.5pt] {};
            }
            \node at (2.5, 3) [above=2mm] {\textbf{Signature}};
            \draw[black, ->] (0, 0) -- (0, 3.03);
            \draw[black, ->] (0, 0) -- (5.03, 0);
            \node at (5.3, 0) {\small $X^1$};
            \node at (0, 3.3) {\small $X^2$};
            \pgfmathsetmacro{\xprev}{\xs[0] + 0.05}
            \pgfmathsetmacro{\yprev}{\ys[0]}
            \foreach \i in \plotrange {
                \pgfmathsetmacro{\x}{\xs[\i]}
                \pgfmathsetmacro{\y}{\ys[\i]}
                \node at (\x, \y)[circle, draw=black, fill=colorp3, inner sep=1.5pt] {};
                \draw[colorp3, thick, cap=round] (\xprev, \yprev) -- (\x, \y);
                \global\let\yprev=\y
                \global\let\xprev=\x
            }
            \node at (3.1, 2.0) [right=0.1mm] {$S^{(1, 2)}$};
            \node at (1.2, 1.1) [right=0.1mm] {$S^{(2, 1)}$};
            
            \draw[black!60, line width=0.3mm, dashed, ->] (\xs[0] + 0.05, \ys[0]) -- (\xs[\xlen] - 0.1, \ys[0]);
            
            \draw[black!60, line width=0.3mm, dashed, ->] (\xs[\xlen], \ys[0]) -- (\xs[\xlen], \ys[\xlen] - 0.1);
            
            \node at (2.5, \ys[0]) [below] {\footnotesize $\Delta X^1$};
            \node at (\xs[\xlen], 1.5)  [right] {\footnotesize $\Delta X^2$};

                        
            \draw[xshift=6cm, name path=one] plot coordinates {
                \myarr
            };
            \draw[xshift=6cm, name path=two] plot coordinates {
                (\xs[0], \ys[0]) (\xs[\xlen], \ys[\xlen])
            };
            \tikzfillbetween[
                of=one and two, split
            ] {pattern=north west lines, colorp2!30};
            \draw[step=0.5, draw=black!30] (6, 0) grid (11, 3) rectangle (6, 0);
            \pgfmathsetmacro{\xprev}{6 + \xs[0] + 0.05}
            \pgfmathsetmacro{\yprev}{\ys[0]}
            
            \draw[black!60, line width=0.3mm, dashed, ->] (\xprev + 0.05, \yprev) -- (\xs[\xlen] + 6 - 0.1, \yprev);
            \draw[black!60, line width=0.3mm, dashed, ->] (\xs[\xlen] + 6, \yprev) -- (\xs[\xlen] + 6, \ys[\xlen] - 0.1);
            \node at (6 + 2.5, \yprev) [below] {\footnotesize $\Delta X^1$};
            \node at (\xs[\xlen] + 6, 1.5)  [right] {\footnotesize $\Delta X^2$};
            \node at (7.1, 1.55) [right=0.1mm] {\small $A_{-}$};
            \node at (9.35, 1.75) [right=0.1mm] {\small $A_{+}$};
            
            \foreach \i in \plotrange {
                \pgfmathsetmacro{\x}{\xs[\i] + 6}
                \pgfmathsetmacro{\y}{\ys[\i]}
                \node at (\x, \y)[circle, draw=black, fill=colorp3, inner sep=1.5pt] {};
                \draw[colorp3, thick, cap=round] (\xprev, \yprev) -- (\x, \y);
                \global\let\yprev=\y
                \global\let\xprev=\x
            }
            \node at (6 + 2.5, 3) [above=2mm] {\textbf{Log-signature}};
            \draw[black, ->] (6, 0) -- (6, 3.03);
            \draw[black, ->] (6, 0) -- (6 + 5.03, 0);
            \node at (6 + 5.3, 0) {\small $X^1$};
            \node at (6, 3.3) {\small $X^2$};
        
        \end{tikzpicture}
    }
    \caption{Geometric depiction of the depth-2 signature and log-signature.  The depth-1 term of both transforms equate to the displacements of the path over the interval in each coordinate, these being $\Delta X^{1, 2}$. \textbf{Left:} The signature. Depth-2 terms $S^{(1, 2)}, S^{(2, 1)}$ correspond to the areas of the blue and orange regions respectively. \textbf{Right:} The log-signature. Only one depth-2 term which is given by the signed area $A_{+} - A_{-}$. This is known as the \textit{L\'evy area} of the path.}
    \label{fig:geometric_signature}
\end{figure*}

	The approach taken by the signature method, coming from rough path theory \citep{levy-lyons, FritzVictoir10}, is to interpret a multivariate time series as a discretisation of an underlying continuous path. The \emph{signature transform}, also known as the \emph{path signature} or \emph{signature}, can then be applied, which produces a vector of real-valued features that are known to characterise the path.
	
	Benefits of the signature method include: a high degree of flexibility, making it possible to customise the method to specific datasets; strong theoretical guarantees; an interpretable feature set; ease of handling irregularly sampled and/or partially observed data; and it being well-defined for some highly irregular processes such as ARMA, Gaussian processes or even Brownian motion. Also, signature features do not need to learned, which can make them particularly effective on (but not limited to) low sample datasets. 
	
	The flexibility of the signature method has made it possible to be tailored to specific applications and achieve state-of-the-art performance in wide range of problem domains, such as handwriting recognition \citep{wilson2018path, yang2016dropsample}, action recognition \cite{yang2016deepwriterid, yang2017leveraging}, and medical time series prediction tasks \citep{morrill2019sepsis, morrill2020utilization}. However, this flexibility comes at the cost of additional complexity in the model search space.
	
	To the best of our knowledge, no comprehensive studies exist that collate and combine the most common method variations found in the literature and assemble them under a common mathematical framework. Additionally, no baseline signature model has ever been tested against other time series classification baselines. Our goal will be to address both of these issues, alongside the development of an open source implementation, so as to make the methods more accessible to a wider audience.

	\paragraph{Contributions}
	We introduce a \emph{generalised signature method} that contains the many existing variations as special cases. In doing so we are able to understand their conceptual groupings into what we term \emph{augmentations}, \emph{windows}, \emph{transforms} and \emph{rescalings}. This involves a comprehensive review of the existing variations across the literature. 
	By understanding their commonality, we are then able to combine different variations, and propose new options that fit into this framework.
    
	We go on to examine which choices within this framework are most important to success by performing an extensive empirical study across 26 datasets. To the best of our knowledge this is the first study of this type. 
	
	In doing so, we are then able to produce a canonical signature pipeline. This represents a domain agnostic starting point that may then be adapted for the task at hand. We show that the performance of this canonical pipeline is comparable to current state-of-the-art classifiers for multivariate time series classification, including deep recurrent and convolutional neural networks. This has led to the implementation of this generalised approach in the open source [redacted] package.
	
	
	\section{Context} \label{sec:context}

	\subsection{Background theory}
	We begin with a few mathematical definitions necessary throughout the article.
    
	\begin{definition} \label{def:mts_space}
	Let $d \in \naturals$, we denote the space of time series over $\reals^d$ as
	\begin{equation*}
	\tseries{\reals^d} = \{(x_1, \ldots, x_n) \,\vert\, x_i \in \reals^d, n \in \naturals, n \geq 1\}.
	\end{equation*}
	\end{definition}
	 If $d=1$, then $\mathbf{x}$ is a univariate time series, whereas if $d >1$, $\mathbf{x}$ is a multivariate time series. Given $\mathbf{x} = (x_1, \ldots, x_n) \in \tseries{\mathbb{R}^d}$, $n$ is called the length of $\mathbf{x}$ and $d$ its dimension or number of channels. We assume that in addition to the array of values $\mathbf{x} \in \tseries{\reals^d}$, we have access to a vector of increasing time stamps $\mathbf{t}=(t_1, \dots, t_n)$. If the data is regularly sampled, then $\mathbf{t}$ can be set to $\mathbf{t}=(1, \dots, n)$, which will often be the case. 
	
	
	We consider a \textit{dataset} to be a collection of such samples. Note that the time stamps $\mathbf{t}$ for each sample may be different, and the sample lengths $n$ can vary. That is, we accept varying length and irregular sampling without modification. We are now in a position to define the signature of a time series.
	
	\begin{definition} \label{def:signature}
	Let $\mathbf{x} \in \tseries{\reals^d}$ and $\mathbf{t}=(t_1, \dots, t_n)$ its associated timestamps. Let $X=(X^1_t, \dots, X^d_t)_{t \in [t_1,t_n]}$ be a piecewise linear interpolation of $\mathbf{x}$ such that for any $i \in \{1, \dots, n\}$, $X_{t_i}=x_i$. Then the depth-$N$ signature transform of $\mathbf{x}$ is the vector defined by
    \begin{align*}
      \mathrm{Sig}^N(\mathbf{x}) = \big( &\{S(\mathbf{x})^{(i)}\}_{i = 1}^{d},\nonumber\\&\{S(\mathbf{x})^{(i, j)}\}_{i, j = 1}^{d},\nonumber\\&\ldots,\nonumber\\&\{S(\mathbf{x})^{(i_1,\ldots, i_N)}\}_{i_1, \ldots, i_N = 1}^{d}\big) \in \reals^{\frac{d^{N+1}-1}{d-1}}
    \end{align*}
    where for any $(i_1, \dots, i_k) \in \{1, \dots, d \}^k$,
    \begin{equation*} \label{eq:sig_depth_N}
        S(\mathbf{x})^{(i_1,\dots,i_k)} = \idotsint\limits_{t_1 \leq u_1 < \dots <u_k \leq t_n} \mathrm{d}X_{u_1}^{i_1} \dots \mathrm{d}X_{u_k}^{i_k} \in \reals.
    \end{equation*}
    \end{definition}
    While this definition may seem somewhat technical, there are several intuitions that can be made with regard to the signature features. We present a geometric interpretation of the first two levels of the signature and log-signature in Figure \ref{fig:geometric_signature}. The depth-1 terms, $S(\mathbf{x})^{(i)}$, equate to the displacement of the path over the interval in the $i$th coordinate, denoted by $\Delta X^i$ in Figure \ref{fig:geometric_signature}. The depth-2 terms, $S(\mathbf{x})^{(i,j)}$, have interpretations in areas generated over the interval.
    
    %
    
    From a statistical point of view, the signature can be thought of as the equivalent of a moment-generating function for time series. Let $Z$ be a random variable, then the moment-generating function of $Z$ is the function 
    \[t \mapsto \mathbb{E}[e^{tz}] = \sum_{k=0}^\infty \frac{t^k}{k!}\mathbb{E}[Z^k]\] and, if well-defined, it characterizes the distribution of $Z$. Assume that $X$ is a random time series (that is a stochastic process), its signature now has the same properties as a moment-generating function: the powers of $Z$ are replaced by integrals of products of coordinates and \citet{chevyrev2016characteristic} show that the expected signature characterizes the law of $X$.
    
    Moreover, we have the following two properties that make the signature a good feature set in a machine-learning context---precise statements may be found in \citet[Appendix A]{kidger2019deep}. 
    \paragraph{Uniqueness} \citet{hambly2010uniqueness} show that under mild assumptions, the full collection of features $\mathrm{Sig}(\mathbf{x})= \lim_{N \to \infty} \mathrm{Sig}^N(\mathbf{x})$ uniquely determines $\mathbf{x}$ up to translations and reparametrizations.
    \vspace{-0.3cm}
    \paragraph{Universal nonlinearity} Linear functionals on the signature are dense in the set of functions on $\mathbf{x}$. Suppose we wish to learn the function $f$ that maps data $\mathbf{x}$ to labels $y$, the universal nonlinearity property states that, under some assumptions, for any $\varepsilon>0$, there exists a linear function $L$ such that
    \begin{equation} \label{eq:universal_nonlinearity}
        \| f(\mathbf{x}) - L\big(\mathrm{Sig}(\mathbf{x})\big)\| \leq \varepsilon.
    \end{equation}

Note that contrary to Fourier or wavelet basis, signatures provide a natural basis for functions of the time series rather than for the time series itself---\eqref{eq:universal_nonlinearity} concerns $f(\mathbf{x})$ and not $\mathbf{x}$. In the context of time series classification this shift of perspective is particularly well-suited since the object of interest is not the time series itself but its link to a label.

	
%
	   
	From a computational point of view, computing the depth-$N$ signature of a time series $\mathbf{x} \in \tseries{\reals^d}$ of length $n$ has a complexity of $\mathcal{O}(n d^N)$, which can be done with high performance software \citep{iisignature, signatory}. The size of the depth-$N$ signature is $(d^{N+1}-1)/(d-1)$ so the memory cost is independent of the series length $n$, which is a huge advantage when dealing with high frequency time series. Note that small values of $N$ already show a good performance---for example $N=3$ in the baseline algorithm---so the exponential dependence on $N$ is not a huge computational bottleneck. 
	
	\paragraph{Logsignature transform} The signature contains some redundant information: for example we can see in the left panel of Figure \ref{fig:geometric_signature} that the sum of the blue and orange areas is equal to the product of displacements $\Delta X^1 \Delta X^2$:
	\begin{equation*}
	    S(\mathbf{x})^{(1, 2)} + S(\mathbf{x})^{(2, 1)} = S(\mathbf{x})^{(1)} S(\mathbf{x})^{(2)}.
	\end{equation*}
	The \textit{logsignature transform} is essentially the signature with these redundancies removed. For example, the logsignature encodes the blue and orange areas from the left panel with the orange signed area in the right panel. However, the logsignature does not have a universal nonlinearity property such as \eqref{eq:universal_nonlinearity}. We refer the reader to \citet{morrill2020logode} or \citet[Section 2]{logsig-rnn} for a precise definition of the logsignature.
	
	A pedagogical introduction to the background theory of signatures is \citet{levy-lyons}, whilst a comprehensive textbook is \citet{FritzVictoir10}. For introductions to the signature method, we recommend \citet[Appendix A]{kidger2019deep} and \citet{primer2016}.
	
	\subsection{Related work} \label{subsec:related_work}
	The signature transform has been used in a wide range of applications in machine learning predictive tasks. For example, as mentioned in the introduction, the signature has been used as a feature extraction layer in classifiers for both Arabic \citep{wilson2018path} and Chinese \citep{yang2016dropsample} handwriting recognition. Similarly, it was successfully used in human action recognition by \citet{li2017lpsnet, yang2017leveraging, logsig-rnn} and in the medical domain as part of the top performing model at the Physionet 2019 challenge for prediction of sepsis \citep{reyna2019early, morrill2019sepsis, morrill2020utilization}. Other applications involve finance \citep{lyons2014feature, perez2018derivatives}, mental health \citep{kormilitzin2017detecting, arribas2018signature}, and emotion recognition \citep{wang2019path, wang2020learning}.
	
	In almost all these applications, the method has been utilised in different ways. Many authors consider transformations of the input time series before application of the signature \citep{levin2013learning, flint2016discretely,lyons2017sketching,yang2017leveraging,logsig-rnn,signatory,invis-reset}. People have also explored different windows over which the signature transform should be taken, so as to extract information over different scales \citep{yang2017leveraging, kidger2019deep}. Additionally a choice must be made between the signature and logsignature transforms, as must choices for the scaling of the terms in the signature \citep{primer2016,normsig}.
	
	The differences between some of these choice have been shown by \citet{fermanian2019embedding} to significantly impact the performance of the methodology. However this study used a small collection of datasets and considered only some of the most common variations that exist in the literature. There is therefore a need for a comprehensive study and unification of all these different choices.

	\section{The generalized signature method}\label{sec:method}	
	In this section we collate the modifications to the signature transform that have been proposed in signature literature to date. We will show that each can be categorised into one of the following groups:
	\begin{itemize}
	    \itemsep0.2em     
	    \item \textbf{Augmentations\quad} These describe the transformation of a time series into one or more new series, in order to return different information in the signature features and deal with dimensionality issues.
	    \item \textbf{Windows\quad} Splitting the time series over different subsequences (or windows), so that signatures may be applied locally.
	    \item \textbf{Transform\quad} The choice between the signature or the logsignature transform.
	    \item \textbf{Rescaling\quad} Ways of normalising the terms in the signature.
	\end{itemize}
	We then go on to show that these groupings can themselves be synergised into a single mathematical framework that we term \textit{the generalised signature method}. For clarity, we will begin by discussing each of these individually, and then afterwards show how they may be combined. 
	
	As before, assume that we observe some collection of sequences $\mathbf{x} \in \tseries{\reals^d}$ with timestamps $\mathbf{t} \in \tseries{\reals}$. 	
	

	

	\subsection{Augmentations}\label{section:augmentations}

	We define an \textit{augmentation} to be a transform of an initial sequence $\mathbf{x} \in \tseries{\reals^d}$ into one or several new sequences. Augmentations have several different uses:
	\begin{enumerate}
	    \item Remove the signature invariance to translation and/or reparametrization.
	    \item Lower the dimension $d$ of the time series, so that higher orders of the signature are reachable---recall that the depth-$N$ signature is of size $\bigO(d^N)$.
	    \item Preprocess the time series prior to the signature map so that information is more easily extracted.
	\end{enumerate}
	
	
	For some $e,p \in \naturals$, we define an augmentation as a map
	\begin{equation*}
	\phi \colon \tseries{\reals^d} \to \tseries{\reals^e}^p.
	\end{equation*}
   There are many pre-signature operations which have been proposed in the literature, and which we categorise as augmentations. We refer the reader to Appendix \ref{sec:augmentations_details} for full details of the many such operations proposed in the literature, but will focus on several important examples here.
   
   Let us give some examples in the first group of sensitivity-inducing augmentations. For any vector of increasing timestamps $\mathbf{t}$, we call \textit{time augmentation} \citep{levin2013learning} the operation $\phi_{\mathbf{t}}: \tseries{\reals^d} \mapsto \tseries{\reals^{d+1}}$ defined by
    \begin{equation} \label{eq:def_time_augmentation}
      \phi_{\mathbf{t}}(\mathbf{x})= \big( (t_1,x_1), \dots, (t_n,x_n) \big).
    \end{equation}
    This transformation, which basically consists in adding the timestamps as an extra coordinate, has two key properties: it guarantees the uniqueness of the signature \citep{hambly2010uniqueness} and it adds information about the parametrization of the time series.
    
    Another example is the basepoint augmentation \citep{signatory}, which is the map $\phi^b: \tseries{\reals^d} \mapsto \tseries{\reals^d}$ defined by
    \begin{equation}
    \label{eq:def_basepoint_augmentation}
	\phi^b(\mathbf{x})= (0,x_1,\dots,x_n),
	\end{equation}
	which simply adds a zero at the beginning of the time series---note that this zero could also be put at the end. This transformation makes the signature sensitive to translations of the time series. The invisibility-reset transformation \citep{yang2017leveraging, wu2020signature} also adds translation sensitivity, but does so by increasing the dimension.
	
	
	In the second group of augmentations for dimensionality reduction, we consider random projections \citep{lyons2017sketching}, which consist in applying multiple random linear maps to the time series, or coordinate projections, which project along (multiple subsets of) the coordinate axes.
	
	In the third group, the lead-lag augmentation \citep{ primer2016,flint2016discretely, yang2017leveraging} captures the quadratic variation by transforming the time series to
	\begin{align*}
	   \phi(\mathbf{x})= \big(&(x_1,x_1),(x_2,x_1),(x_2,x_2),(x_3, x_2), (x_3, x_3), \\
	   &\ldots, (x_n,x_n) \big) \in \tseries{\reals^{2d}}.
	\end{align*}
	Another important example of this kind of augmentation are the stream preserving neural networks of \citet{kidger2019deep}, who learn a map $\phi$ from the data. They map a time series in $\reals^d$ to another series in $\reals^e$ by setting $\phi$ to some neural network, typically either convolutional or recurrent. We extend this idea by defining the multi-headed stream preserving augmentation, which simply consists in stacking $p$ such transformations. To our knowledge, this is the first time that such learned augmentations are compared to `handcrafted' ones such as time, basepoint and lead-lag augmentations.
	
	Importantly, these various augmentations may be combined together. For example, in order to add sensitivity to both parametrization and translation, the time and basepoint augmentations may be combined: first apply the time augmentation, which gives a sequence $\phi_{\mathbf{t}}(x) \in \tseries{\reals^{d+1}}$, and then the basepoint augmentation, which yields
	\begin{equation} \label{eq:composition_augs}
	    \phi^b \circ \phi_{\mathbf{t}}(\mathbf{x})= \big((0,0), (t_1,x_1), \dots, (t_n,x_n) \big).
	\end{equation}
	

	\subsection{Windows}\label{section:windows}
	The second step is to choose a windowing operation. Much like the window functions used with a short time Fourier transform, this localises the signature computation to extract information over particular time intervals.
	
	We define a window to be a map 
	\begin{equation*}
	W \colon \tseries{\reals^e} \to \tseries{\reals^e}^w,
	\end{equation*}
	for some $w \in \naturals$. In short, $W$ maps a time series in $\reals^e$ into $w$ new time series in the same space. The simplest possible window is the global window, defined by
	\begin{equation}\label{eq:globalwindow}
	W(\mathbf{x}) = (\mathbf{x}),
	\end{equation}
	which outputs the time series itself. To get finer-scale information, we consider three other types of windows: sliding, expanding and hierarchical dyadic windows. For $\mathbf{x} =  (x_1, \ldots, x_n) \in \tseries{\reals^e}$ and $ 1 \leq i \leq j \leq n$, let $\mathbf{x}_{i:j} = (x_i, \ldots, x_j) \in \tseries{\reals^e}$ be a subsequence of $\mathbf{x}$. Then, a sliding window of length $\ell$ and step $l$ is defined by
	\begin{equation*}\label{eq:slidingwindow}
	W(\mathbf{x})  = (\mathbf{x}_{1: \ell}, \mathbf{x}_{l + 1: l + \ell}, \mathbf{x}_{2l + 1: 2l + \ell}, \ldots),
	\end{equation*}
	and an expanding window of initial length $\ell$ and step $l$ by
	\begin{equation*}\label{eq:expandingwindow}
	W(\mathbf{x}) = (\mathbf{x}_{1: \ell}, \mathbf{x}_{1: l + \ell}, \mathbf{x}_{1: 2l + \ell}, \ldots).
	\end{equation*}
	The expanding window produces time series of increasing length, and is analogous to the history processes of stochastic analysis whereas the sliding window produces time series of fixed length but shifted in time.
	
	Finally we consider a hierarchical dyadic window, which captures information at different scales. Let $q \in \naturals$ be fixed and assume for simplicity that $2^{q-1}$ divides $n$. Then, the hierarchical dyadic window of depth $q$ consists of $q$ sliding windows $W^1,\dots, W^q$, where $W^i$ has length and step both equal to $n2^{-(i-1)}$. This yields $w=2^{q}-1$ time series of length $n$, $n/2,\, n/4, \, \dots, \, n/2^{q-1}$.  The larger the value of $q$, the finer the scale on which the information is extracted. If the other window functions are analogous to the short time Fourier transform, then hierarchical dyadic windows are analogous to the multi-scale nature of wavelets. 
	
	
	\subsection{The signature and logsignature transforms}
	Central to the signature methodology is of course the signature transform itself. Two choices must be made; whether to use the signature or logsignature transform, and what depth to calculate the transform to---that is, what depth $N$ in Definition \ref{def:signature} to use. Choosing a logsignature lowers the feature vector dimension at the cost of loosing linear approximation properties. There is no consensus on which one should be favored for a machine learning task.
	
	

	\subsection{Rescaling}\label{section:rescaling}
	The depth-$k$ term in the signature is of size $\bigO(\ttfrac{1}{k!})$. Typically, rescaling these terms to $\bigO(1)$ will aid in subsequent learning procedures. To this end, we can apply \textit{pre-signature} scaling whereby we scale the path before signature computation, or \textit{post-signature} where we scale the signature terms themselves. Specifics on how this is done in practice are given in Appendix \ref{sec:rescaling}.
	
	\subsection{Putting the pieces together}
	Let $ \phi \colon \tseries{\reals^d} \to \tseries{\reals^e}^p$ be the final augmentation function, $\phi : \mathbf{x} \mapsto \big( \phi^1(\mathbf{x}), \ldots, \phi^p(\mathbf{x}) \big),$
	which can be a composition of augmentations such as \eqref{eq:composition_augs}. Let $W: \tseries{\reals^e} \to \tseries{\reals^e}^w$, $W : \mathbf{x} \mapsto \big(W^1(\mathbf{x}), \ldots, W^w(\mathbf{x}) \big)$,
	be the window map, such that $W^j(\mathbf{x}) \in \tseries{\reals^e}$ for any $1 \leq j\leq w$. Let $S^N$ represent either the signature or logsignature transform of depth $N$. Let $\rho_{\mathrm{pre}}$ and $\rho_{\mathrm{post}}$ represent the different types of features rescaling. Then given an input $\mathbf{x} \in \tseries{\reals^d}$, the general framework for extracting signature features is given by the collection of
	\begin{equation}\label{eq:generalisedsignaturemethod}
	\mathbf{z}_{i, j} = (\rho_{\mathrm{post}} \circ S^N \circ \rho_{\mathrm{pre}} \circ W^{j} \circ \phi^i)(\mathbf{x})
	\end{equation}
	over all $i \in \{1, \dots, p\}$, $j \in \{1, \dots, w\}$. We refer to the procedure of computing $\mathbf{x} \mapsto (\mathbf{z}_{i, j})$ as the \emph{generalised signature method}.
	
	This final procedure is a little involved, but is simply a combination of different elementary operations used to impact the final feature set. The overall procedure now offers a degree of flexibility and generality which has, to our knowledge, never been achieved for signature methods.

	The collection of features $(\mathbf{z}_{i, j})$ may then be fed into any later machine learning algorithm, which will depend on the application. In general, the $\mathbf{z}_{i,j}$ will be stacked together and considered as a vector. However, if one wants to use a sequential algorithm such as a recurrent network, it is possible to turn the features $\mathbf{z}_{i,j}$ into a sequence by choosing a sliding or expanding window. Indeed, these windows induce an ordering in the features: the terms $\mathbf{z}_{i,1}$ will correspond to the first values of $\mathbf{x}$, the terms $\mathbf{z}_{i,2}$ to the following values, and so on.
	
	
	\section{Empirical study}
	\label{sec:empirical_study}
	We perform a first-of-its-kind empirical study across 26 datasets to determine the most important aspects of this framework.

	\subsection{Methodology}
	\paragraph{Datasets} The datasets used are the Human Activities and Postural Transitions dataset provided by \citet{reyes2016transition}, the Speech Commands dataset provided by \citet{warden2018speech}, and 24 datasets from the UEA time series classification archive, provided by \citet{bagnall2018uea}. A few datasets from the UEA archive were excluded due to their high number of channels resulting in too large a computational burden. 
	
	\paragraph{Baseline} We begin by defining a single baseline procedure, representing a simple and straightforward collection of choices for the generalised signature method. This baseline is to take the augmentation $\phi$ as appending time as defined by \eqref{eq:def_time_augmentation}, $W$ as the global window defined by \eqref{eq:globalwindow}, have the transform be a signature transform of depth $3$, and to use pre-signature scaling of the path. This means that the input features are the collection
	\begin{equation*}
	\mathbf{z}=\mathrm{Sig}^3 \circ \rho_{\mathrm{pre}}\circ \phi_{\mathbf{t}}(\mathbf{x}).
	\end{equation*}
	
	\paragraph{Individual variations} With respect to this baseline procedure, we then consider, in turn, the groups described in Section \ref{sec:method}. These were \textit{augmentations}, \textit{windows}, \textit{transform}, and \textit{rescaling}. For each group we modify the baseline by implementing each option in the group one-by-one. Each such variation defines a particular form of the generalised signature method as in \eqref{eq:generalisedsignaturemethod}. Example variations are to switch to using a logsignature transform of depth 5, or to use a sliding window instead of a global window. We discuss the precise variations below.
	
	\paragraph{Models} On top of every variation, we then consider four different models: logistic regression, random forest, Gated Recurrent Unit (GRU) \citep{gru}, and a residual Convolutional Neural Network (CNN) \citep{resnet}. We test nearly every combination of dataset, variation of the generalised signature method, and model. Different datasets and variations produce different numbers of features $\mathbf{z}_{i, j}$, so to reduce the computational burden we omit those cases for which the number of features is greater than $10^5$. Of the 9984 total combinations of dataset, variation, and model, this leaves out 1415 combinations. See Appendix \ref{sec:omitted_experiments} for a break down of the omitted combinations by different cases.
	
	\paragraph{Analysis} We define the performance of a variation on a dataset as the best performance across the four models considered, to reflect the fact that different models are better suited for different problems. We then follow the methodology of \citet{demvsar2006statistical,benavoli2016should,ruiz2020benchmarking} to compare the variations across the multiple datasets. We first perform a Friedman test to reject the null hypothesis that all methods are equivalent. If it is rejected, we perform pairwise Wilcoxon signed-rank tests to form cliques of not-significant methods, and use critical difference plots to visualize the performance of each signature method. 
	
	A critical difference plot shows the different variations ordered by their average rank: for example, in Figure \ref{fig:basic_augs}, the best variation is ``Time + Basepoint'' with an average rank of 2.5. Then, a thick line indicates that the Wilcoxon test between variations inside the clique is not rejected at significance threshold of 5\%, subject to Bonferroni's multiple testing correction. In Figure \ref{fig:basic_augs} there are two groups of significantly different variations: one with ``Basepoint'' and ``None'' and one with all other variations. 

	We refer the reader to Appendix \ref{sec:implementation_details} for further details on the methodology, such as precise architectural choices, learning rates, and so on.

	\subsection{Results}
	
	Due to the large number of variations and datasets considered, we present only the critical difference plots in the main paper. See Appendix \ref{sec:additional_results} for all the tables of the underlying numerical values.
	
	\begin{figure}[h]
	  	\centering
	  	\vspace{-1em}
  		\includegraphics[width=\linewidth]{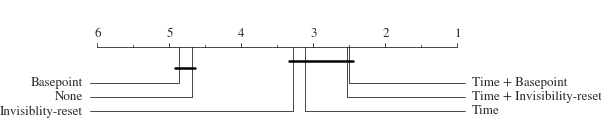}
  		\caption{Performance of invariance-removing augmentations.}
  		\label{fig:basic_augs}
	\end{figure}
	
	\paragraph{Augmentations}
	We split the augmentations into two categories. The first category consists of those augmentations which remove the signature's invariance to translation (basepoint augmentation, invisibility-reset augmentation) or reparameterisation (time augmentation). We see in Figure \ref{fig:basic_augs} that augmenting with time, and either basepoint or invisibility-reset, are both typically important. This is expected; in general a problem need not be invariant to either translation or reparametersiation.
	
	\begin{figure}[h]
	  	\centering
	  	\vspace{-1em}
  		\includegraphics[width=\linewidth]{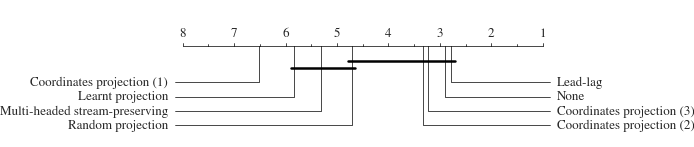}
  		\caption{Performance of other augmentations.}
  		\label{fig:all_augs}
	\end{figure}
	
	The second category consists of those augmentations which either seek to reduce dimensionality or introduce additional information. We see in Figure \ref{fig:all_augs} that most augmentations actually do not help matters, except for lead-lag which usually represents a good choice. We posit that the best augmentation is likely to be dataset dependent, so we break this down by dataset characteristics in Table \ref{tab:augs_ranks_by_type}.
	
	\begin{table}[ht]
    \small
    \setlength{\tabcolsep}{5pt}
	\centering
	\caption{Average ranks for different augmentations by data type. Lower is better. CP (2) stands for coordinate projections with pairs and LP for Learnt projections.}
	\label{tab:augs_ranks_by_type}
	\begin{tabular}{lccccc}
	\toprule
	& \multicolumn{5}{c}{\textbf{Augmentation}}\\
	 \cmidrule{2-6}
	 \textbf{Data type}& None & Lead-lag & CP (2) & LP &  MHSP    \\
	\midrule
	EEG & 4.88 & 4.83 & 3.13 & \textbf{2.75} & \textbf{2.75} \\
	HAR & 2.25 & \textbf{1.78} & 3.50 & 6.50 & 6.50 \\
	MOTION & 2.63 & \textbf{1.75} & 4.50 & 7.33 & 5.00 \\
	OTHER & 2.88 & 3.92 & \textbf{2.63} & 6.00 & 5.21 \\
	\bottomrule           
	\end{tabular}
\end{table}

	Here we indeed see that there is generally a better choice than doing nothing at all, but that this better choice is dependent on some characteristic of the dataset. For example, learnt projections and multi-headed stream preserving transformations do substantially better on EEG datasets, while lead-lag is better for human action and motion recognition.
	
	\begin{figure}[h]
  		\centering
	  	\vspace{-1em}
  		\includegraphics[width=\linewidth]{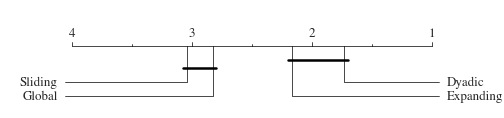}
  		\caption{Performance of different windows.}
  		\label{fig:window}
	\end{figure}
	
	\paragraph{Windows} We consider the possibility of global, sliding, expanding, and dyadic windows. The results are shown in Figure \ref{fig:window}. We see that the dyadic and expanding windows are significantly better than sliding and global windows. The poor performance of sliding windows is a little surprising, but tallies with the observations of \citet{fermanian2019embedding}. This is an important finding, as global and sliding windows tend to be commonly used with signature methods.

	\paragraph{Signature versus logsignature transforms} We consider the signature and logsignature transforms with depths ranging from 1 to 6. As higher depths always produce more information, we define the performance of the (log)signature transform as the best performance across all depths. With this metric, the signature transform is significantly better than the logsignature transform, with a p-value of 0.01 for the Wilcoxon signed-rank test. 


    \paragraph{The key results} To conclude, these results show that invariance-removing transformations such as time and basepoint augmentations should a priori be used, that the lead-lag performs well but not significantly better than no additional augmentation, and that the hierarchical dyadic window performs significantly better than the sliding and global ones. The poor performance of deep learning approaches for augmentation is also notable and an additional motivation for this work: although slightly technical, the augmentations tailored to the signature transform are a significant addition in a machine learning pipeline and cannot be easily replaced by neural networks.
    
    \subsection{Further results}
	See Appendix \ref{sec:additional_results} for further results, in particular on the running times, the different types of rescaling, augmentations broken down by dataset characteristics, an additional study on signature depth, and the precise numerical results for each individual test considered here.

	\section{The canonical signature pipeline}
	\label{sec:performance}
	In this section we define \textit{the canonical signature pipeline}. Using the results from Section \ref{sec:empirical_study} we evaluate the top performing options over all the datasets so as to provide a domain-agnostic starting point for any dataset, from which other variations can be easily explored. We show that this pipeline shows competitive performance against traditional benchmarks and even against deep neural networks.
	
	\subsection{Definition} \label{sec:def_canonical_sig}
	\begin{figure*}[ht]
    \centering
    \resizebox{\linewidth}{!}{
        \begin{tikzpicture}[box/.style={draw=#1, line width=1mm,inner sep=0.8mm}]
            \pgfmathsetseed{\seed}
            \node at (\graphlen/2, \yu) [above=2mm] {\huge \textbf{Input}, $\mathbf{x}$};
            \draw[draw=white, fill=lightgray] (0, -\yd) grid (\graphlen, \yu) rectangle (0, -\yd);
            \origin{0}{0}{$O$}

            \brownian{\numpoints}{\pointdistance}{\randone}{colorp1}{}{\curveshift}{\startone}
            \brownian{\numpoints}{\pointdistance}{\randtwo}{colorp2}{}{\curveshift}{\starttwo}
            \brownian{\numpoints}{\pointdistance}{\randthree}{colorp3}{}{\curveshift}{\startthree}
            
            \pgfmathsetseed{\seed}
            \node at (\Bs + \graphlen/2, \yu) [above=2mm] {\huge \begin{tabular}{c}\textbf{Augmentations, $\phi^b\circ \phi_{\mathbf{t}}$}\\\Large Add time \& basepoint\end{tabular}};
            \draw[draw=white, fill=lightgray] (\Bs, -\yd) grid (\Bs + \graphlen, \yu) rectangle (\Bs, -\yd);
            \brownian{\numpoints}{\pointdistance}{\randone}{colorp1}{}{\curveshift + \Bs}{\startone}
            \brownian{\numpoints}{\pointdistance}{\randtwo}{colorp2}{}{\curveshift+ \Bs}{\starttwo}
            \brownian{\numpoints}{\pointdistance}{\randthree}{colorp3}{}{\curveshift + \Bs}{\startthree}
            \draw[line width=0.5mm, dashed, colorp1] (\Bs, 0) -- (\Bs + \curveshift, \startone);
            \draw[line width=0.5mm, dashed, colorp2] (\Bs, 0) -- (\Bs + \curveshift, \starttwo);
            \draw[line width=0.5mm, dashed, colorp3] (\Bs, 0) -- (\Bs + \curveshift, \startthree);
            \draw[line width=0.5mm, dashed, black] (\Bs, 0) -- ((\Bs + \graphlen, \yu) node[pos=1,sloped, anchor = north east] {\large Time};
            \node at (\Bs, 0) [circle,fill,inner sep=2pt]{};
            \draw[pen colour=black, decoration={calligraphic brace, amplitude=5pt, mirror}, black, decorate, line width=1pt] (\Bs, -1) -- (\Bs + 1, -1) node[below, midway, yshift=-0.7, font=\large] {Basepoint} ;
            
            \pgfmathsetseed{\seed}
            \node at (\Cs + \graphlen/2, \yu) [above=2mm] {\huge \begin{tabular}{c}\textbf{Window, $W^j$}\\\Large Hierarchical dyadic, with $j$ optimised.\end{tabular}};
            \def\yshift{0}
            \foreach\i in {0}{
                \pgfmathsetseed{\seed}  
                \def\xstart{\Cs + \i*\graphlen/4}
                \def\ymid{3*\yu/4-\yshift}
                \def\ytop{\ymid+\yu/4}
                \def\ybottom{\ymid-\yu/4}
                \draw[draw=white, fill=lightgray] (\xstart, \ybottom) grid (\xstart + \graphlen/4, \ytop) rectangle (\xstart, \ybottom);
                \draw[line width=0.5mm, black] (\xstart, \ymid) -- ((\xstart + \graphlen/4, \ytop) node[below right] {};
                \brownian{\numpoints}{0.34 * \pointdistance}{\randone/4}{colorp1}{}{\curveshift/4 + \xstart}{\startone/4+\ymid}
                \brownian{\numpoints}{0.34 * \pointdistance}{\randtwo/4}{colorp2}{}{\curveshift/4 + \xstart}{\starttwo/4+\ymid}
                \brownian{\numpoints}{0.34 * \pointdistance}{\randthree/4}{colorp3}{}{\curveshift/4 + \xstart}{\startthree/4+\ymid}
                \draw[line width=0.5mm, colorp1] (\xstart, \ymid) -- (\xstart + \curveshift/4, \startone/4+\ymid);
                \draw[line width=0.5mm, colorp2] (\xstart, \ymid) -- (\xstart + \curveshift/4, \starttwo/4+\ymid);
                \draw[line width=0.5mm, colorp3] (\xstart, \ymid) -- (\xstart + \curveshift/4, \startthree/4+\ymid);
                \node [draw=colord1, dashed, fill=none, line width=0.8mm, minimum height=2cm, minimum width=2cm] at (\xstart + \graphlen/8, \ymid) {};
            }
            \def\yshift{3}
            \foreach\i in {0, 1}{
                \pgfmathsetseed{\seed}  
                \def\xstart{\Cs + \i*\graphlen/4}
                \def\ymid{3*\yu/4-\yshift}
                \def\ytop{\ymid+\yu/4}
                \def\ybottom{\ymid-\yu/4}
                \draw[draw=white, fill=lightgray] (\xstart, \ybottom) grid (\xstart + \graphlen/4, \ytop) rectangle (\xstart, \ybottom);
                \draw[line width=0.5mm, black] (\xstart, \ymid) -- ((\xstart + \graphlen/4, \ytop) node[below right] {};
                \brownian{\numpoints}{0.34 * \pointdistance}{\randone/4}{colorp1}{}{\curveshift/4 + \xstart}{\startone/4+\ymid}
                \brownian{\numpoints}{0.34 * \pointdistance}{\randtwo/4}{colorp2}{}{\curveshift/4 + \xstart}{\starttwo/4+\ymid}
                \brownian{\numpoints}{0.34 * \pointdistance}{\randthree/4}{colorp3}{}{\curveshift/4 + \xstart}{\startthree/4+\ymid}
                \draw[line width=0.5mm, colorp1] (\xstart, \ymid) -- (\xstart + \curveshift/4, \startone/4+\ymid);
                \draw[line width=0.5mm, colorp2] (\xstart, \ymid) -- (\xstart + \curveshift/4, \starttwo/4+\ymid);
                \draw[line width=0.5mm, colorp3] (\xstart, \ymid) -- (\xstart + \curveshift/4, \startthree/4+\ymid);
                \node [draw=colord2, dashed, fill=none, line width=0.8mm, minimum height=2cm, minimum width=1cm] at (\xstart + \graphlen/16 + \i*\graphlen/8, \ymid) {};
            }
            \def\yshift{0}
            \foreach\i in {0, 1, 2, 3}{
                \pgfmathsetseed{\seed}  
                \def\xstart{\Cs + \i*\graphlen/4}
                \def\ymid{-3*\yu/4}
                \def\ytop{\ymid+\yu/4}
                \def\ybottom{\ymid-\yu/4}
                \draw[draw=white, fill=lightgray] (\xstart, \ybottom) grid (\xstart + \graphlen/4, \ytop) rectangle (\xstart, \ybottom);
                \draw[line width=0.5mm, black] (\xstart, \ymid) -- ((\xstart + \graphlen/4, \ytop) node[below right] {};
                \brownian{\numpoints}{0.34 * \pointdistance}{\randone/4}{colorp1}{}{\curveshift/4 + \xstart}{\startone/4+\ymid}
                \brownian{\numpoints}{0.34 * \pointdistance}{\randtwo/4}{colorp2}{}{\curveshift/4 + \xstart}{\starttwo/4+\ymid}
                \brownian{\numpoints}{0.34 * \pointdistance}{\randthree/4}{colorp3}{}{\curveshift/4 + \xstart}{\startthree/4+\ymid}
                \draw[line width=0.5mm, colorp1] (\xstart, \ymid) -- (\xstart + \curveshift/4, \startone/4+\ymid);
                \draw[line width=0.5mm, colorp2] (\xstart, \ymid) -- (\xstart + \curveshift/4, \starttwo/4+\ymid);
                \draw[line width=0.5mm, colorp3] (\xstart, \ymid) -- (\xstart + \curveshift/4, \startthree/4+\ymid);
                \node [draw=colord3, dashed, fill=none, line width=0.8mm, minimum height=2cm, minimum width=0.5cm] at (\xstart + \graphlen/32 + \i*\graphlen/16, \ymid) {};
            }
            \node at (\Ds + \graphlen/2, \yu) [above=2mm] {\huge \begin{tabular}{c}\textbf{Transform, $S^N$}\\\Large Signature features, with $N$ optimised.\\\end{tabular}};
            \path[fill=lightgray] (\Ds, -\yd) rectangle (\Ds + \graphlen, \yu);
            \def\st{\Ds +  2.5*\xgap}
            \def\ygap{0.95}
            \def\xgap{1}
            \def\ngap{0.5}
            \def\csize{8pt}
            \def\egap{1.5}
            \def\nyu{\yu-0.2}
            \foreach \x in {1, 2, 3} {
                \filldraw[colord1] (\st + \xgap*\x, \nyu - \ngap) circle (\csize);
                \node[colord1] at (\st + \xgap*3 + \egap, \nyu - \ngap) {\Huge\textbf{\ldots}};
                \filldraw[colord1] (\st + \xgap*\x + 3.18*\egap, \nyu - \ngap) circle (\csize);
                \foreach \y in {1, 2} {
                    \filldraw[colord2] (\st + \xgap*\x, \nyu - \y*\ygap - 2*\ngap) circle (\csize);
                    \ifthenelse{\x=1}
                        {
                            \node[colord2] at (\st + \xgap*3 + \egap, \nyu - \y*\ygap - 2*\ngap) {\Huge\textbf{\ldots}};
                        }{}
                    \filldraw[colord2] (\st + \xgap*\x + 3.18*\egap, \nyu - \y*\ygap - 2*\ngap) circle (\csize);

                }
                \foreach \y in {3, 4, 5, 6} {
                    \filldraw[colord3] (\st + \xgap*\x, \nyu - \y*\ygap - 3*\ngap) circle (\csize);
                    \ifthenelse{\x=1}
                        {
                            \node[colord3] at (\st + \xgap*3 + \egap, \nyu - \y*\ygap - 3*\ngap) {\Huge\textbf{\ldots}};
                        }{}
                    \filldraw[colord3] (\st + \xgap*\x + 3.18*\egap, \nyu - \y*\ygap - 3*\ngap) circle (\csize);
                }
            }
            
            \node at (\Es + \graphlen/2, \yu + 1) [above=2mm] {\Large Stack};
            \draw[fill=lightgray, lightgray] (\Es + \graphlen/2 - 0.5, -5) rectangle (\Es + \graphlen/2 + 0.5, 5);
            \def\eps{0.5}
            \foreach \y in {0, 1, 2} {
                \filldraw[colord1] (\Es + \graphlen/2, \yu + \eps -  \y*\ygap) circle (\csize);
                \filldraw[colord3] (\Es + \graphlen/2, -\yd - \eps + \y*\ygap) circle (\csize);
            }
            \def\finalposup{\yu + \eps -  2*\ygap}
            \def\finalposdown{-\yd - \eps + 2*\ygap}
            \def\finalposgap{\finalposup - \finalposdown}
            \node[colord1, rotate=90] at (\Es + \graphlen/2, \finalposup - 1) {\huge\textbf{\ldots}};
            \node[colord2, rotate=90] at (\Es + \graphlen/2, \finalposdown - 0.5 * \finalposgap + 0.65) {\huge\textbf{\ldots}};
            \node[colord3, rotate=90] at (\Es + \graphlen/2, \finalposdown + 1) {\huge\textbf{\ldots}};
            \draw[thick, line width=0.5mm] (\Ds + \graphlen + 3 * \xgap - 0.3, \nyu - 6*\ygap - 3*\ngap) to [square right brace] (\Ds + \graphlen + 3 * \xgap  - 0.3, \nyu-\ngap);
            
            \draw[->, line width=0.5mm] (\graphlen + 0.02, 0) -- (\Bs + 0.85 * \xgap, 0);
            \draw[->, line width=0.5mm] (\Bs + \graphlen + \xgap + 0.02, 0) -- (\Cs + 2.8 * 0.65 * \xgap, 0);
            \draw[->, line width=0.5mm] (\Cs + \graphlen + 2 * \xgap + 0.05, 0) -- (\Ds + 3 * \xgap,0);
            \draw[-, line width=0.5mm] (\Ds + \graphlen + 3 * \xgap + 0.2, 0) -- (\Ds + \graphlen + 4.5 * \xgap, 0);
            \draw[->, line width=0.5mm] (\Ds + \graphlen + 5.5 * \xgap, 0) -- (\Ds + \graphlen + 6.5 * \xgap, 0);
            

            \draw[fill=lightgray, draw=none] (\Ds + \graphlen + 6.5 * \xgap, -2) rectangle (\Ds + \graphlen + 6.5 * \xgap + \graphlen/1.5, 2) node[pos=.5] {\huge\begin{tabular}{c}Black box\\ML classifier\end{tabular}};
        \end{tikzpicture}
    }
    \caption{Pictorial representation of the canonical signature pipeline. First, we apply the time and basepoint augmentations to the input paths, then we compute the signature features over dyadic windows, and finally compute the signature features over each dyadic window. These features can now be compiled together and fed into any standard machine learning classifier.}
    \label{fig:gensig}
\end{figure*}
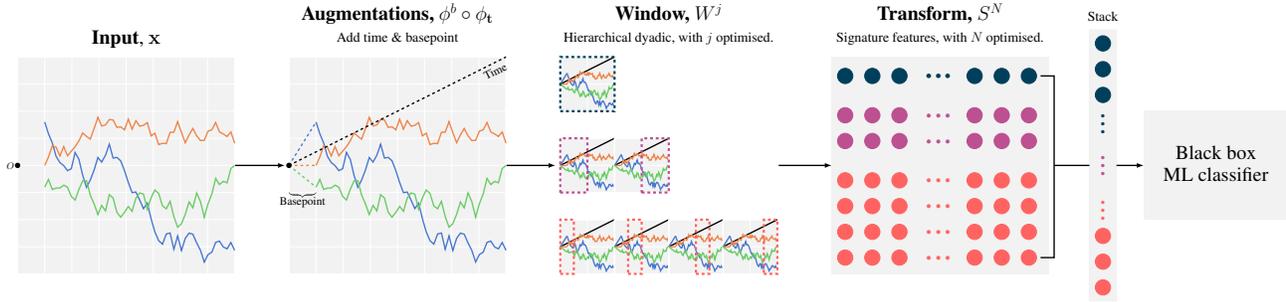
    In a nutshell, the pipeline consists in applying the basepoint and time augmentations, a hierarchical dyadic window and a signature transform, which can be written as a particular case of \eqref{eq:generalisedsignaturemethod} as follows. Let $W$ be a hierarchical dyadic window of depth $q$, $\phi_{\mathbf{t}}$ and $\phi^b$ be the time and basepoint augmentations, then the canonical signature pipeline may be written as
    \begin{equation} \label{eq:canonical_sig_pipeline}
        \mathbf{z}_j=S^N \circ W^j \circ \phi^b \circ \phi_{\mathbf{t}}(\mathbf{x}), \quad j \in \{1, \dots, 2^q-1\}.
    \end{equation}
	We give a graphical depiction of this in Figure \ref{fig:gensig}. Signature and window depths ($N$, $q$) must be optimised for the problem (typically via cross-validation). We note that this canonical method may be adapted to the problem at hand in two ways: if the problem is known to be parametrization invariant, as is the case for example for characters recognition, then the time augmentation should not be applied. Moreover, if the problem is translation-invariant, then the basepoint augmentation is not applied. We emphasise that this pipeline does not represent a best option for every application, but is meant to represent a compromise between broad applicability, ease of implementation, computational cost, and good performance. 

	\subsection{Performance}
    
    We validate the performance of the pipeline against the 26 datasets in the multivariate UEA archive\footnote{This is not to be confused with the UCR archive which is a collection of 128 univariate datasets.}. To our knowledge, the most recent benchmarks for the UEA archive are the results from \citet{ruiz2020benchmarking}. We compare their results to the canonical signature pipeline with a random forest classifier---see Appendix \ref{sec:best_rf_details} for more details.


    The benchmarks include variants on classical Dynamic Time Wrapping (DTWI, DTWD and DTWA); an ensemble of univariate classifiers, HIVE COTE \citep{bagnall2020tale}, known to be highly perfromant in the univariate case; a random shapelet forest \citep{karlsson2016generalized}, denoted gRSF, and a bag of words based algorithm, MUSE \citep{schafer2017fast}; two deep learning methods, TapNet \citep{wang2017time} and MLCN \citep{karim2019multivariate}. The MLCN architecture combines long short term memory layers (LSTM) and convolutional layers while TapNet combines 3 blocks: random projections on the different dimensions, convolutional layers and a final attention block to compare candidate time series representations.
    \begin{figure}[h]
  		\centering
  		\includegraphics[width=\linewidth]{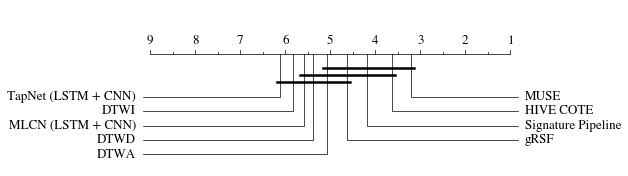}
  		\caption{Performance on UEA datasets.}
  		\label{fig:best_rf}
	\end{figure}
	
	Figure \ref{fig:best_rf} shows the critical difference plot of this comparison. The signature pipeline is in the first clique, that is the group of classifiers that achieve the best accuracy while not being significantly different from one another. The two algorithms with a better rank than the signature pipeline are MUSE and HIVE-COTE. It is worth noting that MUSE is very memory intensive---\citet{ruiz2020benchmarking} report that it could not finish on 5 of the 26 UEA datasets on a computer with 500GB of memory---whilst HIVE-COTE is an ensemble of several sub-classifiers, and thus has very high training and inference costs. On the other hand, all experiments for the canonical signature pipeline were completed with no memory errors on a computer with less memory, and are significantly faster to run than HIVE-COTE---see Appendix \ref{sec:additional_results}. 
	
	
	The canonical signature pipeline is meant to be a sensible starting point from which the user can propose additional variations following the structure defined in \eqref{eq:generalisedsignaturemethod}, but as a standalone classifier this pipeline performs comparably to state-of-the-art classifiers, on the UEA data, whilst being less computationally demanding.
	
	
	\section{Conclusion}
	We introduce a generalised signature method as a framework to capture recently proposed variations on the signature method. We go on to perform a first-of-its-kind extensive empirical investigation as to which elements of this framework are most important for performance in a domain-independent setting. In particular, we highlight the performance of hierarchical dyadic windows and signature-tailored augmentations such as lead-lag, time and basepoint. As a result, we are able to present a canonical signature pipeline that represents a best-practices domain-agnostic starting point, which shows competitive performance against state-of-the-art classifiers for multivariate time series classification.
    
	

\bibliography{references}
\bibliographystyle{style/icml2021}

	\normalsize
	\newpage
	\onecolumn
\appendix
\begin{center}
\huge Supplementary material
\end{center}

\section{Augmentations}
	\label{sec:augmentations_details}

	We recall that an augmentation is a map
	\begin{align*}
	\phi \colon  \tseries{\reals^d} &\to \tseries{\reals^e}^p
	\end{align*}
	We give below the precise definition of the different augmentations considered in the study, which are summarized in Table \ref{tab:summary_augmentations}. These augmentations were not typically introduced using such language, so this serves as a reference for how the existing literature may be interpreted through the generalised signature method.	
	
	Throughout the section, we consider a sequence $\mathbf{x}=(x_1,\dots,x_n) \in \tseries{\reals^d}$ and timestamps $\mathbf{t}=(t_1, \dots, t_n) \in \tseries{\reals}$. We recall that if $\mathbf{x}$ is regularly sampled then $\mathbf{t}$ is usually set to $\mathbf{t}=(1, \dots, n)$.

	\paragraph{Time augmentation}
	

    We recall the definition of the time augmentation:
	\begin{equation*}
	\phi(\mathbf{x})= \big((t_1,x_1),\dots,(t_n,x_n) \big) \in \tseries{\reals^{d+1}}.
	\end{equation*}
	It ensures uniqueness of the signature transformation and removes the parametrization invariance \citep{levin2013learning}.

	\paragraph{Invisibility-reset augmentation}

	First introduced by \citet{yang2017leveraging}, the invisibility-reset augmentation consists in adding a coordinate to the sequence $\mathbf{x}$ that is constant equal to 1 but drops to 0 at the last time step, i.e.,
	\begin{equation*}
	\phi(\mathbf{x})= \big((1,x_1),\dots,(1,x_{n-1}) ,(1,x_n),(0,x_n),(0,0) \big) \in \tseries{\reals^{d+1}}.
	\end{equation*}
	This augmentation adds information on the initial position of the path, which is otherwise not included in the signature as it is a translation-invariant map.

	\paragraph{Basepoint augmentation}
	Introduced by \citet{signatory}, the basepoint augmentation has the same goal as the invisibility-reset augmentation: removing the translation-invariant property of the signature. It simply adds the point 0 at the beginning of the sequence:
	\begin{equation*}
	\phi(\mathbf{x})= (0,x_1,\dots,x_n) \in \tseries{\reals^{d}}.
	\end{equation*}
	The main difference compared to the invisibility-reset augmentation is that the signature of $\mathbf{x}$ is contained in the signature of the invisibility-reset augmented path, whereas it is not in the signature of the basepoint augmented path. The price paid is that the invisibility-reset augmentation introduces redundancy into the signature, and is more computationally expensive due to the additional channel. (Recall that the signature method scales as $\bigO(d^N)$, where $d$ is the input channels and $N$ is the depth of the (log)signature.)

	\paragraph{Lead-lag augmentation}

	The lead-lag augmentation, introduced by \citet{primer2016} and \citet{flint2016discretely} has been used in several applications (see for example \citet{lyons2014feature,kormilitzin2016application,yang2017leveraging}). It adds lagged copies of the path as new coordinates. This then explicitly captures the quadratic variation of the underlying process \citep{flint2016discretely}. As many different lags as desired may be added. If there is a single lag of a single timestep, then this corresponds to
	\begin{equation*}
	\phi(\mathbf{x})= ((x_1,x_1),(x_2,x_1),(x_2,x_2),\dots,(x_n,x_n)) \in \tseries{\reals^{2d}}.
	\end{equation*}
	
	\begin{table}
    \centering
    \caption{Summary of the different augmentations}
    \label{tab:summary_augmentations}
    \begin{tabular}{llll l}
        \toprule
         & $e$ & $p$ & Property  \\
        \midrule
        \multicolumn{4}{l}{Fixed augmentations} \\
        \cmidrule(r){1-1}
        None                                & $d$   & 1     &      \\
        Time                                & $d+1$ & 1     & sensitivity to parametrization, \\
        									&		&		& uniqueness of the signature map \\
        Invisibility-reset                  & $d+1$ & 1     & sensitivity to translation  \\
        Basepoint                           & $d$   & 1     & sensitivity to translation  \\
        Lead-lag   	                        & $2d$  & 1     & information about quadratic variation, \\
        									&		&		& uniqueness of the signature map \\
        Coordinates projection &     &   & dimensionality reduction \\
        \qquad with singletons & 2 & $d$ & \\
        \qquad with pairs  & 3     & $d(d-1)$ &  \\
        \qquad with triplets  & 4     & $d(d^2-1)$&\\
        Random projections                  & $e$   & $p$     & dimensionality reduction \\
        \cmidrule(r){1-1}
        \multicolumn{4}{l}{Learnt augmentations} \\
        \cmidrule(r){1-1}
        Learnt projections 	& $e$	& $p$		& data-dependent and linear \\
        Stream-preserving neural network & $e$ & $1$ & data-dependent \\
        Multi-headed stream-preserving NN      & $e$   & $p$   &  data-dependent  \\
        \bottomrule
    \end{tabular}
\end{table}

	\paragraph{Coordinate projections}
	For multidimensional streams, one may want to compute the signature of a subset of coordinates individually, rather than the signature of the whole stream; doing so restricts the interaction considered by the signature to just those between the projected coordinates. Let $\mathbf{x^1},\dots, \mathbf{x^d} \in \tseries{\reals}$ denote the different coordinates of $\mathbf{x}\in \tseries{\reals^d}$.
	
	Then we define the singleton coordinate projection as
	\begin{equation*}
	\phi(\mathbf{x})= \big((\mathbf{t},\mathbf{x^1}),(\mathbf{t},\mathbf{x^2}),\dots,(\mathbf{t},\mathbf{x^{d}}) \big) \in \tseries{\reals^2}^{d},
	\end{equation*}
	whilst considering all possible pairs of coordinates yields the augmentation
	\begin{equation*}
	\phi(\mathbf{x})= \big((\mathbf{t}, \mathbf{x^1},\mathbf{x^2}),(\mathbf{t}, \mathbf{x^1},\mathbf{x^3}),\dots,(\mathbf{t}, \mathbf{x^d},\mathbf{x^{d-1}}) \big) \in \tseries{\reals^3}^{d(d-1)},
	\end{equation*}
	and all possible triples yields the augmentation
	\begin{equation*}
	\phi(\mathbf{x})= \big((\mathbf{t}, \mathbf{x^1},\mathbf{x^1},\mathbf{x^2}),(\mathbf{t}, \mathbf{x^1},\mathbf{x^1},\mathbf{x^3}),\dots,(\mathbf{t}, \mathbf{x^{d}},\mathbf{x^d},\mathbf{x^{d-1}}) \big) \in \tseries{\reals^4}^{d(d^2-1)}.
	\end{equation*}
	
	The decision to always include a time dimension is a somewhat arbitrary one, and it may alternatively be excluded if desired. (This is done so as to make sense of singleton coordinate projections; otherwise the result is a collection of univariate time series, for which the signature extracts only the increment due to the tree-like equivalence property.)
	
	\paragraph{Random projections}
	When the dimension of the input path is very large, \citet{lyons2017sketching} have proposed to project it into a smaller space by taking multiple random projections. Let $e<d$ and let $A_i:\reals^d \to \reals^{e}$ be random affine transformations indexed by $i \in \{1, \ldots, p\}$. Then $\phi$ is defined as
	\begin{equation*}
	\phi(\mathbf{x})= ((A_1 x_1,\dots,A_1 x_n), \ldots, (A_p x_1,\dots,A_p x_n)) \in \tseries{\reals^{e}}^p.
	\end{equation*}
	
	\paragraph{Learnt projections}
	Rather than taking random projections, \citet{logsig-rnn} learn it from the data. This takes exactly the same form as the random projections, except that the $A_i$ are learnt.
	
	\paragraph{Stream-preserving neural network}
	\citet{kidger2019deep} introduce arbitrary learnt sequence-to-sequences maps prior to the signature transform, and refer to such maps, when parameterised as neural networks, as stream-preserving neural networks. For example these may be standard convolutional or recurrent architectures. In general this may be any learnt transformation
	\begin{equation*}
	\phi \colon \tseries{\reals^d} \to \tseries{\reals^e}.
	\end{equation*}

	\paragraph{Multi-headed stream-preserving neural network}
	A straightforward extension of stream-preserving neural networks is to use multiple such networks, so as to avoid a potential bottleneck through the single signature map that it is eventually used in. Letting $\phi^1, \ldots, \phi^p$ be $p$ different stream-preserving neural networks, then this gives an augmentation
	\begin{equation*}
	\phi(\mathbf{x}) = (\phi^1(\mathbf{x}), \ldots, \phi^p(\mathbf{x})) \in (\tseries{\reals^e})^p.
	\end{equation*}
	
	\section{Rescaling}\label{sec:rescaling}
	The signature transform can be written as a sequence of tensors, indexed by $k \in \{1, \ldots, N\}$. The $k$-th term is of size $\bigO(\ttfrac{1}{k!})$, as it is computed by an integral over a $k$-dimensional simplex. It is typical that rescaling these terms to be $\bigO(1)$ will aid subsequent learning procedures. 
	
	One option is to simply multiply the $k$-th term by $k!$, which we call \textit{post-signature} scaling.
	
	However, it is possible that the previous option may suffer from numerical stability issues. Thus we also explore the performance of an option, called \textit{pre-signature scaling}, which may alleviate this, which is to multiply the input $\mathbf{x}$ by some scaling factor $\alpha \in \reals$. Then the $k$-th term will be of size $\bigO(\ttfrac{\alpha^k}{k!})$, and so by taking $\alpha = (N!)^{\ttfrac{1}{N}}$ the $N$-th term in the signature will be $\bigO(1)$; the trade-off is that Stirling's approximation then shows that the $\ttfrac{N}{2}$-th term will be of size $\bigO(2^{N/2})$.

	\section{Implementation details}
	\label{sec:implementation_details}
	
	\subsection{General notes}
	
	\paragraph{Code} All the code for this project is available at [redacted for anonymity].
	
	\paragraph{Libraries} The machine learning framework used was PyTorch \citep{pytorch} version 1.3.1. Signatures and logsignatures were computed using the Signatory library \citep{signatory} version 1.1.6. Scikit-learn \citep{scikit-learn} version 0.22.1 was used for the logistic regression and random forest models. The experiments were tracked using the Sacred framework \citep{greff2017sacred} version 0.8.1.

	\paragraph{Normalisation} Every dataset was normalised so that each channel has mean zero and unit variance.
	
	\paragraph{Architectures} Two different GRU models were used on every dataset; a `small' one with 32 hidden channels and 2 layers, and a `large' one with 256 hidden channels and 3 layers.
	
	Likewise, two different Residual CNN models were considered. The `small' one used 6 blocks, each composed of batch normalisation, ReLU activation, convolution with 32 filters and kernel size 4, batch normalisation, ReLU activation, and a final convolution with 32 filters and kernel size 4, so that there are also 32 channels along the `residual path'. A final two-hidden-layer neural network with 256 neurons was placed on the output. The `large' is similar, except that it used 128 filters in both the blocks and the residual path, had 8 blocks, used a kernel size of 8, and the final neural network had 1024 neurons.
	
	The logistic regression was performed three times with different amounts of $L^2$ regularisation, with scaling hyperparameters of 0.01, 0.2 and 1; for every experiment the regularization hyperparameter achieving the best accuracy on the test set was used.
	
	The random forest used the default Scikit-learn implementation with a maximum depth of 6 and 100 trees.
	
	\paragraph{Optimiser}The GRU and CNN were optimised using Adam \citep{adam}. The learning rate was 0.01 for the GRU, and 0.001 for the residual CNN. The small models were trained for a maximum of 500 epochs; the large models were trained for a maximum of 1000 epochs. The learning rate was decreased by a factor of 10 if validation loss did not improve over a plateau of 10 epochs. Early stopping was used if the validation loss did not improve for 30 epochs. After training the parameters were always rolled back to those that demonstrated the best validation loss over the course of training. The batch size used varied by dataset; in each it was taken to be the power of two that meant that the number of batches per epoch was closest to 40.
	
	\paragraph{Computing infrastructure}Experiments were run on an Amazon AWS G3 Instance (g3.16xlarge) equipped with 4 Tesla M60s, parallelized using GNUParallel \citep{Tange2011a}.

	\subsection{Analysis of variations of the signature method}\label{sec:omitted_experiments}\label{sec:experiment_details}
	\paragraph{Splits}
	The UEA archive comes with a pre-defined train-test split, which we respect. We take an 80\%/20\% train/validation split in the training data, stratified by class label. For the Human Activities and Postural Transitions dataset, we take a 60\%/15\%/25\% train/validation/test split from the whole dataset. For the Speech Commands dataset, we take a 68\%/17\%/15\% train/validation/test split from the whole dataset. (These somewhat odd choices corresponding to taking either 25\% or 15\% of the dataset as test, and then splitting the remaining 80\%/20\% between train and validation.) These train/validation splits are only used for the training of the GRU and CNN classifiers.
	
	\paragraph{Combinations}
	In total we tested 8569 different combinations.
	
	The variations tested are divided into groups. The first group consists of the sensitivity-adding augmentations, namely time, basepoint and invisibility-reset. Relative to the baseline model, we test every possible combination of these. (Including using none of them.)
	
	The second group consists of those other augmentations, namely the lead-lag, singleton coordinate projection, pair coordinate projection, triplet coordinate projection, random projections, learnt projections, and multi-headed stream preserving neural networks, and finally also the case of no additional augmentation.
	
	For the random projections, we consider four possibilities, with $e \in \{3, 6\}$ and $p \in \{2, 5\}$, all relative to the baseline model.
	
	For no additional augmentation, lead-lag, coordinate projections, learnt projections, and the multi-headed stream preserving neural networks, we compose them with the time, time+basepoint and time+invisibility-reset augmentations (the clear best three from the first group), all relative to the baseline model.
	
	For the learnt projections, we consider four different possibilities corresponding to $e \in \{3, 6\}$ and $p \in \{2, 5\}$; together with the time/time+basepoint/time+invisibility-reset cases this yields a total of twelve possibilities.
	
	For the multi-headed stream-preserving neural networks, we again consider four different possibilities corresponding to $e \in \{3, 6\}$ and $p \in \{2, 5\}$, for a total of twelve possible augmentation strategies. In each the neural network operates elementwise, so as to map one sequence to another, and is given by a feedforward neural network of three hidden layers separated by ReLU activation functions. When $e = 3$ the hidden layers have 16 neurons each, and when $e = 6$ they have 32 neurons each.
	
	For both the learnt projections and multi-headed stream-preserving neural networks, training these requires backpropagating through the model, so these were only considered for the GRU and residual CNN model. (The logistic regression model would in principle be possible as well, except that we ended up implementing this through Scikit-learn rather than PyTorch.)
	
	We note that there are a great many possible ways of doing stream preserving neural networks, of which these are a small fraction. Their relatively weak performance here may likely be improved upon with greater tuning on an individual task, or the selection of better final models than were considered here.
	
	The third group consisted of the different windows. Recall that the baseline model used a global window; we then consider varying this to two possible sliding windows, two possible expanding windows, and three possible dyadic windows. The two possible sliding/expanding windows are chosen so that either 5 or 20 windows are applied across the full length of the dataset. The three possible dyadic windows are depths 2, 3, 4. Thus in total there are 8 possible window combinations we consider.
	
	The fourth group consists of rescaling options, namely no rescaling, pre-signature rescaling, and post-signature rescaling.
	
	\paragraph{Omissions}
	For the empirical study on the variations on the signature method, we excluded those UEA datasets with a dimension $d$ over 60, so as to reduce the computational cost. This results removes 6 of the 30 datasets from the study, namely DuckDuckGeese, FaceDetection, Heartbeat, InsectWingbeat, MotorImagery, and PEMS-SF. These were nonetheless used in the demonstration of performance of the canonical signature method in Figure \ref{fig:best_rf}. Furthermore those combinations of dataset/variation/model which produced more than $10^5$ signature features were omitted, to keep the computation managable. See Table \ref{tab:omissions}.

	\begin{table}
    \caption{Summary of the number of combinations considered and omitted.}\label{tab:omissions}
    \begin{tabular}{lcccc}
        \toprule
        \textbf{Variations } & \textbf{\# Variations} & \textbf{\# Classifiers} & \textbf{\# Omitted } &\textbf{\# Total}  \\
        & & & \textbf{Combinations} & \textbf{Combinations} \\
        \midrule
        Basic augmentations (Figure \ref{fig:basic_augs}) & $6$ & $6$ & $54$ &  $936$\\
        \cmidrule{1-1}
        Other augmentations (Figure \ref{fig:all_augs}) \\
        Lead-lag/ None & $3$ & $6$ & $100$/$27$ & $468$ \\
        Coordinates projection (1)/(2)/(3)& $3$ & $6$ & $12$/$12$/$54$ & $468$ \\
        Random projections & $4$ & $6$ & $32$ & $624$ \\
        Learnt projections / MHSP & $12$ & $4$ & $348$/$176$ & $1248$\\
        \cmidrule{1-1}
        Windows (Figure \ref{fig:window}) & $8$ & $6$ & $227$ & $1248$ \\
        \cmidrule{1-1}
        Signature/Logsignature transform & $12$ & $6$ &$361$ &$1872$ \\ 
        \cmidrule{1-1}
        Rescalings (Figure \ref{fig:rescaling}) & $3$ & $6$ &$12$ & $468$ \\
        \midrule{}
        \textbf{Total} & & & $1415$ & $9984$ \\
        \bottomrule
    \end{tabular}
\end{table}

	\subsection{The canonical signature pipeline}
	\label{sec:best_rf_details}

	For each dataset, we implement the following steps. First, the sequences are augmented with time and basepoint augmentations. Then, we consider every combination of signature depth in $\{1, 2, 3, 4, 5, 6\}$ and hierarchical dyadic window depth in $\{2, 3, 4\}$. For each of these choices, we perform a randomized grid search on a random forest classifier to optimize its number of trees and maximal depth parameters. We test 20 combinations randomly sampled from the following grids:
	\begin{align*}
	    \textnormal{n\_trees}&=[50, 100, 500, 1000], \\
       \textnormal{max\_depth}&=[2, 4, 6, 8, 12, 16, 24, 32, 45, 60, 80, \textnormal{None}].
    \end{align*}
    Note that a maximal depth set to `None' means that the trees are expanded until all leaves contain exactly one sample. Finally, we choose the combination of signature and hierarchical dyadic window depths which maximise the out-of-bag score. 
     
    \section{Additional results}
	\label{sec:additional_results}

	\subsection{Analysis of variations of the signature method}
	
	\begin{table}
    \caption{Average run time (in seconds) for various experiments. mean (std), averaged over all UEA datasets.}
    \label{tab:average_run_time}
    \centering
    \begin{tabular}{lcccc}
        \toprule
        &\multicolumn{4}{c}{\textbf{Classifier}} \\
        \cmidrule{2-5}
         & \multirow{2}{*}{CNN} & \multirow{2}{*}{GRU} & Logistic  & Random \\
         & & & regression & forest \\
        \midrule
        Time augment \& Global window (Baseline) & 69.8 (98.0) & 22.2 (31.8) & 2.67 (7.09) & 2.23 (4.84) \\
        \midrule
        \textbf{Augmentation} & & & \\
        \cmidrule{1-1}
        None & 48.1 (63.5) & 16.8 (33.6) & 3.55 (9.91) & 66.3 (321) \\
        Lead-lag & 48.58 (69.99) & 15.2 (18.1) & 5.76 (11.7) & 3.35 (6.04) \\
        Coordinates projection (1) & 32.8 (31.49) & {13.4 (17.8) }& 1.37 (4.2) & 12.2 (59.3) \\
        Coordinates projection (2) & 41.5 (51.4) & 22.6 (62.3) & 3.01 (8.54) & 42.3 (203) \\
        Coordinates projection (3) & {41.3 (39.9)} & 19.1 (24.5) & 5.41 (9.76) & 6.3 (14.1) \\
        Random projection & 62.2 (70.1) & 21.1 (31.2) & {0.86 (1.25) }& {1.4 (2.47) }\\
        Learnt projection & 917 (1288) & 752 (972) & -- & -- \\
        Multi-headed stream-preserving & 1051 (1677) & 1758 (4442) & -- & -- \\
        \cmidrule{1-1}
        \textbf{Window}  & & & \\
        \cmidrule{1-1}
        Sliding & 90.6 (120) & 79.4 (175) & 10.1 (27.4) & 6.4 (16.0) \\
        Expanding & 102 (133) & 68.7 (115) & 9.98 (27.2) & 7.17 (19.0) \\
        Dyadic & 725 (868) & 56.9 (65.1) & 12.5 (33.2) & 7.59 (18.3) \\
        \bottomrule
    \end{tabular}
\end{table}
	
	\paragraph{Running time}
	To get a sense of the cost of each augmentation or window, we present the run times of each augmentation/model combination, and each window/model combination. (The times for varying between signature and logsignature, and between different rescalings, are largely insignificant.) See Table \ref{tab:average_run_time}.
	
	The run times are averaged over every UEA dataset. As the datasets are of very different sizes this thus represents quite a crude statistic, and in particular produces very large variances, so these are most meaningful simply with respect to each other.

	\begin{table}[ht]
	\caption{Average ranks for different augmentations by type of data. Lower is better.}
	\label{tab:basic_augs_ranks_by_type}
    \centering
    \begin{tabular}{lllllll}
        \toprule
        & \multicolumn{6}{c}{\textbf{Augmentation}}\\
        \cmidrule{2-7}
    	\multirow{2}{*}{\textbf{Data type}} &\multirow{2}{*}{None } &\multirow{2}{*}{Time} &\multirow{2}{*}{Basepoint}& \multirow{2}{*}{Invisibility-reset}    & Time + & Time +        \\
		& & & & & Basepoint & Invisibility-reset  \\
		\midrule
		EEG & 3.88 & 3.50 & 4.00 & \textbf{2.00} & 4.00 & 3.63 \\
		HAR & 5.00 & 2.95 & 4.85 & 3.65 & \textbf{2.00} & 2.55 \\
		MOTION & 5.25 & 2.75 & 5.75 & 3.88 & \textbf{1.50} & 1.88 \\
		OTHER & 4.43 & 3.31 & 4.88 & 3.19 & 2.87 & \textbf{2.31 } \\ 
		\bottomrule
	\end{tabular}
\end{table}
	
	\paragraph{Sensitivity-inducing augmentations broken down by dataset type}

	Table \ref{tab:basic_augs_ranks_by_type} shows the average rank of each of the first group of augmentations (that add sensitivity to certain kinds of perturbation) by dataset type, where the types are taken from \citet{bagnall2018uea}. (This may be regarded as a companion to Table \ref{tab:augs_ranks_by_type}.)
	
	It is interesting to note that for EEG data, it seems better not to consider the time augmentation, whereas it is the case for other applications. In particular the combination of time and basepoint augmentations achieve the best ranks for human action and motion recognition (HAR and MOTION in Table \ref{tab:basic_augs_ranks_by_type}). Recognizing an action may not be translation-invariant nor invariant by time reparametrization.
	
	\begin{table*}[ht]
    \small
	\centering
	\caption{Average ranks for different augmentations by dataset characteristics. Lower is better.}
	\label{tab:augs_ranks_by_type}
	\begin{tabular}{lcccccccc}
	\toprule
	& \multicolumn{8}{c}{\textbf{Augmentation}}\\
	 \cmidrule{2-9}
	 \multirow{2}{*}{\textbf{Characteristic}}&\multirow{2}{*}{None}& \multirow{2}{*}{Lead-lag}& \multicolumn{3}{c}{Coordinates projection} & \multirow{2}{*}{\begin{tabular}{c}Random\\Projection\end{tabular}} & \multirow{2}{*}{\begin{tabular}{c}Learnt\\Projection\end{tabular}} &  \multirow{2}{*}{MHSP}    \\
	 & & & (1) & (2) & (3) &  &  & \\
	\midrule
	\textbf{Data type} \\
	\cmidrule{1-1}
	EEG & 4.88 & 4.83 & 6.50 & 3.13 & 5.67 & 4.38 & \textbf{2.75} & \textbf{2.75} \\
	HAR & 2.25 & \textbf{1.78} & 7.20 & 3.50 & 2.90 & 4.75 & 6.50 & 6.50 \\
	MOTION & 2.63 & \textbf{1.75} & 7.00 & 4.50 & 2.13 & 5.00 & 7.33 & 5.00 \\
	OTHER & 2.88 & 3.92 & 5.44 & \textbf{2.63} & 3.29 & 4.69 & 6.00 & 5.21 \\
	\midrule
	\textbf{Series length}\\
	\cmidrule{1-1}
	$<$50 & 3.20 & \textbf{2.20} & 7.40 & 3.20 & 2.70 & 5.10 & 7.00 & 5.20 \\
	50-100 & 2.20 & \textbf{1.33} & 6.00 & 4.10 & 2.75 & 6.20 & 4.80 & 5.10 \\
	100-500 & \textbf{2.28} & 2.57 & 7.28 & 3.50 & 2.63 & 4.17 & 6.63 & 5.33 \\
	$>$500 & 4.00 & 4.00 & 5.28 & \textbf{2.64} & 4.57 & 4.07 & 4.40 & 5.60 \\
	\midrule
	\textbf{Dimension $d$}\\
	\cmidrule{1-1}
	2 & 4.67 & 3.5 & 6.33 & 4.33 & 4.0 & \textbf{2.83} & 6.67 & 3.67 \\
	3-5 & 2.5 & \textbf{2.21} & 6.36 & 3.43 & 3.14 & 4.64 & 6.67 & 6.83 \\
	6-8 & 3.25 & \textbf{2.5} & 6.94 & 3.0 & 3.56 & 5.0 & 5.29 & 6.0 \\
	$>$8 & \textbf{2.25} & 3.75 & 6.31 & 3.19 & 2.5 & 5.19 & 5.29 & 4.19\\
	\bottomrule           
	\end{tabular}
\end{table*}

	\paragraph{Other augmentations broken down by dataset characteristichs}
    
    Table \ref{tab:augs_ranks_by_type}presents the average ranks of the other augmentations borken down by some characteristics of the datasets.

	Here we see that there is generally a better choice than doing nothing at all, but that this better choice. For example on long or high-dimensional datasets, coordinate projections often perform well, whilst multi-headed stream preserving transformations do substantially better on EEG datasets. Lead-lag remains a strong choice in many cases.
	
\begin{figure}[h]
  		\centering
  		\includegraphics[width=.6\linewidth]{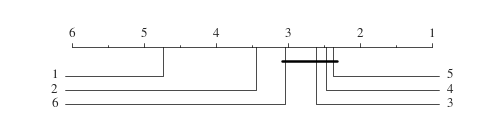}
  		\caption{Critical differences plot for the depth study on the UEA datasets.}
  		\label{fig:depth}
	\end{figure}

	\paragraph{Depth study on the signature transform}
	In the main text we focused on the difference between the signature and logsignature transforms, and stated that larger depths must be chosen by a bias-variance tradeoff. Here we consider varying the depth together with the choice of signature or logsignature, and taking the best transform for each depth. See Figure \ref{fig:depth}. We see that larger depths do indeed generally correspond to increased performance, up to a point. The optimal depth will depend on the complexity of the task, as the number of features increases exponentially with the depth.
	
\begin{figure}[h]
  		\centering
	  	\vspace{-1em}
  		\includegraphics[width=.6\linewidth]{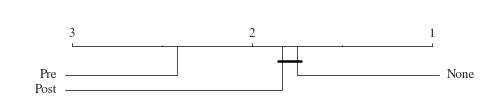}
  		\caption{Performance of different rescalings}
  		\label{fig:rescaling}
	\end{figure}

	\paragraph{Rescaling critical difference diagram} In Figure \ref{fig:rescaling}, we see that pre-signature rescaling performs significantly worse than the other two options and that no significant difference between post-rescaling and no rescaling is found. 

	\subsection{Complete results}

	We present in Tables \ref{tab:complete_results_basic_augs}, \ref{tab:complete_results_all_augs}, \ref{tab:complete_results_windows}, \ref{tab:complete_results_sig_logsig}, \ref{tab:complete_results_rescaling} and \ref{tab:complete_results_depth} the performance of the different signature variations on each dataset. The tables were obtained by maximizing the test accuracy of the signature method over the different classifiers considered. Recall that some values are omitted due to the large number of signature features that would be obtained.

    \begin{table}
    \scriptsize
    \centering
    \caption{Accuracy of sensitivity-inducing augmentations per dataset}
    \label{tab:complete_results_basic_augs}
    \begin{tabular}{lcccccc}
        \toprule
         & \multicolumn{6}{c}{\textbf{Augmentation}}\\
         \cmidrule{2-7}
        \multirow{2}{*}{\textbf{Dataset}} &\multirow{2}{*}{None } &\multirow{2}{*}{Time} &\multirow{2}{*}{Basepoint}& \multirow{2}{*}{Invisibility-reset}    & Time + & Time +        \\
        	& & & & & Basepoint & Invisibility-reset  \\
        \midrule
        ArticularyWordRecognition & 96.0 & 96.3 & 95.7 & 96.3 & 97.7 & 97.0 \\
        AtrialFibrillation & 46.7 & 46.7 & 40.0 & 33.3 & 40.0 & 40.0 \\
        BasicMotions & 100.0 & 100.0 & 100.0 & 100.0 & 100.0 & 100.0 \\
        CharacterTrajectories & 88.3 & 93.2 & 86.4 & 88.7 & 93.8 & 93.7 \\
        Cricket & 91.7 & 94.4 & 94.4 & 97.2 & 97.2 & 95.8 \\
        ERing & 80.0 & 92.6 & 77.4 & 89.6 & 91.9 & 92.2 \\
        EigenWorms & 72.5 & 79.4 & 74.8 & 76.3 & 87.0 & 81.7 \\
        Epilepsy & 84.8 & 89.9 & 91.3 & 91.3 & 97.1 & 94.9 \\
        EthanolConcentration & 27.8 & 29.3 & 33.5 & 41.8 & 34.6 & 41.4 \\
        FingerMovements & 55.0 & 52.0 & 57.0 & 58.0 & 55.0 & 56.0 \\
        HandMovementDirection & 29.7 & 33.8 & 32.4 & 33.8 & 36.5 & 32.4 \\
        Handwriting & 21.9 & 30.8 & 23.6 & 24.6 & 30.6 & 28.7 \\
        JapaneseVowels & 85.4 & 85.1 & 97.3 & 98.1 & 97.3 & 98.1 \\
        LSST & 42.0 & 47.4 & 44.0 & 44.4 & 50.9 & 48.7 \\
        Libras & 72.8 & 84.4 & 65.0 & 75.0 & 80.0 & 77.2 \\
        NATOPS & 81.7 & 88.3 & 79.4 & 79.4 & 91.1 & 92.2 \\
        PenDigits & 91.1 & 97.1 & 88.3 & 93.1 & 96.8 & 97.1 \\
        PhonemeSpectra & 4.7 & 8.2 & 4.3 & 5.7 & 10.0 & 8.1 \\
        RacketSports & 78.9 & 80.3 & 78.9 & 82.9 & 82.9 & 81.6 \\
        SelfRegulationSCP1 & 81.6 & 83.3 & 76.8 & 84.0 & 75.4 & 85.0 \\
        SelfRegulationSCP2 & 57.2 & 56.7 & 56.1 & 56.7 & 56.1 & 55.0 \\
        SpokenArabicDigits & 82.5 & 85.5 & 80.5 & 88.0 & 85.1 & 90.1 \\
        StandWalkJump & 60.0 & 46.7 & 40.0 & 46.7 & 40.0 & 46.7 \\
        UWaveGestureLibrary & 84.1 & 87.5 & 79.7 & 82.8 & 87.5 & 83.4 \\
        Human Activity & 73.0 & 76.6 & 92.3 & 92.2 & 93.0 & 93.8 \\
        Speech Commands & 71.4 & 75.9 & 74.7 & 74.9 & 79.7 & 79.5 \\
        \midrule
        Average rank &  4.69 & 3.12 & 4.87 & 3.29 & \textbf{2.5} & 2.54 \\
        \bottomrule
    \end{tabular}
\end{table}
    \begin{table}
    \scriptsize
    \centering
    \caption{Accuracy of other augmentations per dataset}
    \label{tab:complete_results_all_augs}
    \begin{tabular}{lcccccccc}
        \toprule
        & \multicolumn{8}{c}{\textbf{Augmentation}}\\
         \cmidrule{2-9}
         \multirow{2}{*}{\textbf{Dataset}}&\multirow{2}{*}{None}& \multirow{2}{*}{Lead-lag}& \multicolumn{3}{c}{Coordinates projection} & Random& Learnt &  \multirow{2}{*}{MHSP}    \\
         & & & (1) & (2) &(3) & projection &  projection \\
        \midrule
        ArticularyWordRecognition & 97.7 & 96.3 & 83.3 & 95.7 & 97.0 & 95.3 & 73.7 & 80.3 \\
        AtrialFibrillation & 46.7 & 40.0 & 53.3 & 53.3 & 46.7 & 66.7 & 46.7 & 53.3 \\
        BasicMotions & 100.0 & 100.0 & 80.0 & 100.0 & 100.0 & 100.0 & 97.5 & 87.5 \\
        CharacterTrajectories & 93.8 & 95.3 & 43.9 & 93.2 & 93.8 & 93.3 & 89.6 & 91.1 \\
        Cricket & 97.2 & 98.6 & 90.3 & 97.2 & 95.8 & 88.9 & 69.4 & 56.9 \\
        ERing & 92.6 & 94.8 & 79.3 & 89.3 & 91.9 & 74.4 & 62.2 & 61.1 \\
        EigenWorms & 87.0 & 87.8 & 50.4 & 84.0 & 89.3 & 78.6 & -- & -- \\
        Epilepsy & 97.1 & 97.1 & 55.8 & 95.7 & 95.7 & 81.9 & 67.4 & 65.9 \\
        EthanolConcentration & 41.4 & 39.9 & 42.2 & 43.3 & 42.2 & 30.4 & 32.3 & 30.0 \\
        FingerMovements & 56.0 & -- & 59.0 & 58.0 & -- & 55.0 & 60.0 & 65.0 \\
        HandMovementDirection & 36.5 & 31.1 & 31.1 & 40.5 & 37.8 & 37.8 & 33.8 & 44.6 \\
        Handwriting & 30.8 & 33.5 & 11.3 & 27.8 & 30.0 & 21.6 & 12.6 & 13.2 \\
        JapaneseVowels & 98.1 & 97.6 & 94.1 & 97.8 & 97.6 & 84.1 & 95.4 & 95.9 \\
        LSST & 50.9 & 55.6 & 43.5 & 51.7 & 52.8 & 43.7 & 34.4 & 39.8 \\
        Libras & 84.4 & 86.7 & 47.8 & 83.9 & 85.0 & 86.7 & 73.3 & 81.1 \\
        NATOPS & 92.2 & -- & 33.3 & 90.6 & 91.1 & 85.6 & 83.9 & 81.7 \\
        PenDigits & 97.1 & 98.3 & 60.2 & 96.8 & 97.2 & 96.7 & 96.5 & 97.4 \\
        PhonemeSpectra & 10.0 & -- & 4.5 & 9.4 & 10.6 & 8.9 & 7.2 & 7.7 \\
        RacketSports & 82.9 & 82.2 & 53.3 & 85.5 & 84.2 & 75.7 & 73.7 & 75.0 \\
        SelfRegulationSCP1 & 85.0 & 86.0 & 61.8 & 85.0 & 84.0 & 81.6 & 86.7 & 84.6 \\
        SelfRegulationSCP2 & 56.7 & 57.2 & 55.6 & 58.9 & 55.6 & 60.6 & 59.4 & 57.8 \\
        SpokenArabicDigits & 90.1 & 96.6 & 58.5 & 86.0 & 90.0 & 83.0 & 88.0 & 86.0 \\
        StandWalkJump & 46.7 & 40.0 & 40.0 & 53.3 & 40.0 & 53.3 & -- & -- \\
        UWaveGestureLibrary & 87.5 & 88.8 & 50.6 & 85.6 & 87.5 & 86.2 & 74.1 & 75.6 \\
        Human Activity & 93.8 & 93.6 & 75.8 & 93.2 & 93.6 & 69.2 & 91.3 & 91.5 \\
        Speech Commands & 79.7 & -- & 14.9 & 77.1 & -- & 70.2 & -- & 76.1 \\
        \midrule
        Average ranks & 2.9 & \textbf{2.77} & 6.52 & 3.33 & 3.23 & 4.71 & 5.83 & 5.31\\
        \bottomrule
    \end{tabular}
\end{table}
    \begin{table}
    \scriptsize
    \centering
    \caption{Accuracy of windows per dataset}
    \label{tab:complete_results_windows}
    \begin{tabular}{lcccc}
        \toprule
        & \multicolumn{4}{c}{\textbf{Window}}\\
         \cmidrule{2-5}
        \textbf{Dataset} & Global & Sliding & Expanding & Dyadic  \\
        \midrule
        ArticularyWordRecognition & 96.3 & 89.3 & 99.0 & 99.0 \\
        AtrialFibrillation & 46.7 & 46.7 & 46.7 & 60.0 \\
        BasicMotions & 100.0 & 100.0 & 100.0 & 100.0 \\
        CharacterTrajectories & 93.2 & 94.6 & 96.9 & 97.1 \\
        Cricket & 97.2 & 93.1 & 97.2 & 95.8 \\
        ERing & 90.7 & 88.5 & 91.9 & 94.8 \\
        EigenWorms & 80.2 & 74.8 & 78.6 & 76.3 \\
        Epilepsy & 89.9 & 92.8 & 92.0 & 94.2 \\
        EthanolConcentration & 30.4 & 38.8 & 30.0 & 35.7 \\
        FingerMovements & 50.0 & -- & -- & -- \\
        HandMovementDirection & 33.8 & 33.8 & 36.5 & 33.8 \\
        Handwriting & 30.1 & 21.5 & 30.2 & 27.2 \\
        JapaneseVowels & 85.1 & 76.5 & 88.1 & 89.2 \\
        LSST & 47.8 & 43.1 & 48.3 & 46.6 \\
        Libras & 83.3 & 85.6 & 91.1 & 90.0 \\
        NATOPS & 93.3 & 84.4 & 90.6 & -- \\
        PenDigits & 95.8 & -- & -- & 97.6 \\
        PhonemeSpectra & 8.6 & 9.2 & 9.6 & 10.4 \\
        RacketSports & 82.2 & 80.3 & 84.2 & 88.2 \\
        SelfRegulationSCP1 & 82.6 & 87.4 & 84.0 & 86.7 \\
        SelfRegulationSCP2 & 56.7 & 60.6 & 54.4 & 56.1 \\
        SpokenArabicDigits & 85.5 & 91.5 & 93.7 & 96.6 \\
        StandWalkJump & 46.7 & 53.3 & 46.7 & 53.3 \\
        UWaveGestureLibrary & 86.6 & 79.4 & 89.1 & 89.7 \\
        Human Activity & 76.1 & 73.0 & 80.4 & 81.7 \\
        Speech Commands & 75.9 & 76.5 & 82.3 & 83.0 \\
        \midrule
        Average ranks & 2.83 & 3.04 & 2.17 & \textbf{1.73} \\
        \toprule
    \end{tabular}
\end{table}
    \begin{table}
    \scriptsize
    \centering
    \caption{Accuracy of signature and logsignature transforms per dataset.}
    \label{tab:complete_results_sig_logsig}
    \begin{tabular}{lcc}
        \toprule
        & \multicolumn{2}{c}{\textbf{Transform}}\\
        \cmidrule{2-3}
        \textbf{Dataset}& Signature & Logsignature  \\
        \midrule
        ArticularyWordRecognition & 97.7 & 97.3 \\
        AtrialFibrillation & 60.0 & 53.3 \\
        BasicMotions & 100.0 & 100.0 \\
        CharacterTrajectories & 93.8 & 93.8 \\
        Cricket & 100.0 & 100.0 \\
        ERing & 90.0 & 89.3 \\
        EigenWorms & 79.4 & 81.7 \\
        Epilepsy & 93.5 & 91.3 \\
        EthanolConcentration & 31.9 & 30.0 \\
        FingerMovements & 59.0 & 56.0 \\
        HandMovementDirection & 40.5 & 40.5 \\
        Handwriting & 35.3 & 24.5 \\
        JapaneseVowels & 85.9 & 86.8 \\
        LSST & 52.0 & 46.4 \\
        Libras & 90.6 & 87.8 \\
        NATOPS & 89.4 & 91.7 \\
        PenDigits & 97.8 & 97.5 \\
        PhonemeSpectra & 8.9 & 7.6 \\
        RacketSports & 85.5 & 84.9 \\
        SelfRegulationSCP1 & 84.0 & 83.3 \\
        SelfRegulationSCP2 & 57.2 & 56.1 \\
        SpokenArabicDigits & 87.5 & 85.8 \\
        StandWalkJump & 53.3 & 53.3 \\
        UWaveGestureLibrary & 90.0 & 86.9 \\
        Human Activity & 78.7 & 78.3 \\
        Speech Commands & 75.9 & 76.3 \\
        \midrule
        Average ranks & \textbf{1.25} & 1.75 \\
    \bottomrule
    \end{tabular}
\end{table}
    \begin{table}
	\scriptsize
	\centering
	\caption{Accuracy of rescaling choices per dataset}
	\label{tab:complete_results_rescaling}
	\begin{tabular}{lccc}
	\toprule
	& \multicolumn{3}{c}{\textbf{Rescaling}}\\
	 \cmidrule{2-4}
	\textbf{Dataset}& None & Post & Pre \\
	\midrule
	ArticularyWordRecognition & 97.3 & 97.0 & 97.7 \\
	AtrialFibrillation & 53.3 & 53.3 & 46.7 \\
	BasicMotions & 100.0 & 100.0 & 100.0 \\
	CharacterTrajectories & 94.6 & 94.6 & 94.6 \\
	Cricket & 98.6 & 97.2 & 97.2 \\
	ERing & 93.7 & 93.7 & 93.0 \\
	EigenWorms & 80.9 & 80.9 & 79.4 \\
	Epilepsy & 92.0 & 92.0 & 91.3 \\
	EthanolConcentration & 31.2 & 31.6 & 30.0 \\
	FingerMovements & 54.0 & 54.0 & 50.0 \\
	HandMovementDirection & 35.1 & 32.4 & 29.7 \\
	Handwriting & 36.6 & 36.4 & 37.1 \\
	JapaneseVowels & 87.3 & 85.9 & 85.7 \\
	LSST & 55.8 & 55.6 & 55.4 \\
	Libras & 85.0 & 86.1 & 84.4 \\
	NATOPS & 92.8 & 92.8 & 91.7 \\
	PenDigits & 96.6 & 96.7 & 96.7 \\
	PhonemeSpectra & 8.0 & 8.1 & 8.2 \\
	RacketSports & 84.2 & 84.2 & 83.6 \\
	SelfRegulationSCP1 & 79.5 & 83.3 & 84.6 \\
	SelfRegulationSCP2 & 56.1 & 57.2 & 56.7 \\
	SpokenArabicDigits & 90.5 & 90.5 & 90.2 \\
	StandWalkJump & 46.7 & 53.3 & 46.7 \\
	UWaveGestureLibrary & 87.5 & 87.2 & 87.2 \\
	Human Activity & 85.0 & 84.6 & 85.1 \\
	Speech Commands & 77.0 & 75.7 & 75.9 \\
	\midrule
	Average ranks & \textbf{1.73} & 1.92 & 2.35 \\
	\bottomrule
	\end{tabular}
	\end{table}

	\begin{table}
	\scriptsize
	\centering
	\caption{Accuracy of (log)signature depth per dataset.}
	\label{tab:complete_results_depth}
	\begin{tabular}{lcccccc}
	\toprule
	& \multicolumn{6}{c}{\textbf{Depth}}\\
	 \cmidrule{2-7}
	\textbf{Dataset} & 1 & 2 & 3 & 4 & 5 & 6 \\
	\midrule
	ArticularyWordRecognition & 83.3 & 96.0 & 97.3 & 97.7 & 95.3 & -- \\
	AtrialFibrillation & 40.0 & 40.0 & 60.0 & 33.3 & 40.0 & 53.3 \\
	BasicMotions & 70.0 & 100.0 & 100.0 & 100.0 & 100.0 & 92.5 \\
	CharacterTrajectories & 42.3 & 88.0 & 93.2 & 93.8 & 92.9 & 93.8 \\
	Cricket & 30.6 & 93.1 & 97.2 & 98.6 & 100.0 & -- \\
	ERing & 77.0 & 89.6 & 90.0 & 89.3 & 88.9 & 84.8 \\
	EigenWorms & 46.6 & 81.7 & 79.4 & 79.4 & -- & -- \\
	Epilepsy & 50.7 & 78.3 & 89.9 & 93.5 & 93.5 & 93.5 \\
	EthanolConcentration & 25.5 & 30.8 & 30.0 & 31.2 & 31.9 & 27.4 \\
	FingerMovements & 57.0 & 58.0 & 59.0 & -- & -- & -- \\
	HandMovementDirection & 40.5 & 36.5 & 37.8 & 39.2 & 32.4 & -- \\
	Handwriting & 7.3 & 22.4 & 32.4 & 33.3 & 35.3 & 32.7 \\
	JapaneseVowels & 78.9 & 85.9 & 86.8 & 84.3 & 81.4 & -- \\
	LSST & 40.9 & 45.6 & 47.6 & 50.6 & 52.0 & 44.7 \\
	Libras & 51.7 & 77.2 & 85.0 & 87.8 & 88.9 & 90.6 \\
	NATOPS & 35.0 & 86.7 & 91.7 & -- & -- & -- \\
	PenDigits & 60.0 & 90.4 & 96.9 & 97.7 & 97.4 & 97.8 \\
	PhonemeSpectra & 4.1 & 7.6 & 8.9 & -- & -- & -- \\
	RacketSports & 44.1 & 77.0 & 78.9 & 84.9 & 85.5 & 82.2 \\
	SelfRegulationSCP1 & 53.6 & 80.2 & 84.0 & 83.3 & 81.9 & -- \\
	SelfRegulationSCP2 & 56.1 & 55.0 & 56.7 & 54.4 & 57.2 & -- \\
	SpokenArabicDigits & 52.1 & 85.8 & 85.5 & 87.5 & -- & -- \\
	StandWalkJump & 46.7 & 46.7 & 46.7 & 46.7 & 53.3 & 46.7 \\
	UWaveGestureLibrary & 49.4 & 83.1 & 86.6 & 87.8 & 90.0 & 88.1 \\
	Human Activity & 47.7 & 78.3 & 76.0 & 78.7 & 78.6 & -- \\
	Speech Commands & 14.8 & 69.6 & 76.3 & -- & -- & -- \\
	\midrule
	Average ranks & 4.73 & 3.44 & 2.62 & 2.48 & \textbf{2.38} & 3.04 \\
	\bottomrule
	\end{tabular}
	\end{table}

	\subsection{Canonical signature method}
	In Table \ref{tab:all_results_best_rf} we give the full results for our canonical signature method on all UEA datasets, together with the results of \citet{ruiz2020benchmarking} used in Figure \ref{fig:best_rf}.
	
	\begin{table}
	\scriptsize
	\centering
	\begin{tabular}{lccccccccc}
        \toprule
        & \multicolumn{9}{c}{\textbf{Classification method}}\\
        \cmidrule{2-10}
        \textbf{Dataset} &  DTWD &   DTWA &   DTWI &  HIVE COTE &  MLCN &   MUSE &  TapNet &   gRSF &  \parbox{15mm}{Signature Pipeline}\\
        \midrule
        ArticularyWordRecognition &       98.7 &   98.7 &   98.0 &       99.0 &  95.7 &   99.3 &    95.7 &   98.3 &                97.7 \\
        AtrialFibrillation        &       20.0 &   26.7 &   26.7 &       13.3 &  33.3 &   40.0 &    20.0 &   26.7 &                46.7 \\
        BasicMotions              &       97.5 &  100.0 &  100.0 &      100.0 &  87.5 &  100.0 &   100.0 &  100.0 &               100.0 \\
        Cricket                   &      100.0 &  100.0 &   98.6 &       98.6 &  91.7 &   98.6 &   100.0 &   98.6 &                95.8 \\
        Epilepsy                  &       96.4 &   97.8 &   97.8 &      100.0 &  73.2 &   99.3 &    95.7 &   97.8 &                95.7 \\
        EthanolConcentration      &       32.3 &   31.6 &   30.4 &       79.1 &  37.3 &   47.5 &    30.8 &   34.6 &                43.3 \\
        ERing                     &       91.5 &   92.6 &   91.9 &       97.0 &  94.1 &   97.4 &    90.4 &   95.2 &                94.8 \\
        FaceDetection             &       52.9 &   52.8 &   51.3 &       65.6 &  55.5 &   63.1 &    60.3 &   54.8 &                61.4 \\
        FingerMovements           &       53.0 &   51.0 &   52.0 &       55.0 &  58.0 &   55.0 &    47.0 &   58.0 &                52.0 \\
        HandMovementDirection     &       18.9 &   20.3 &   29.7 &       44.6 &  52.7 &   36.5 &    33.8 &   41.9 &                20.3 \\
        Handwriting               &       60.7 &   60.7 &   50.9 &       48.2 &  30.9 &   52.2 &    28.1 &   37.5 &                37.9 \\
        Heartbeat                 &       71.7 &   69.3 &   65.9 &       72.2 &  38.0 &   71.2 &    79.0 &   76.1 &                69.8 \\
        Libras                    &       87.2 &   88.3 &   89.4 &       90.0 &  85.0 &   89.4 &    87.8 &   69.4 &                93.9 \\
        LSST                      &       55.1 &   56.7 &   57.5 &       57.5 &  52.8 &   64.0 &    51.3 &   58.8 &                56.9 \\
        NATOPS                    &       88.3 &   88.3 &   85.0 &       88.9 &  90.0 &   90.6 &    81.1 &   84.4 &                92.2 \\
        PenDigits                 &       97.7 &   97.7 &   93.9 &       93.4 &  97.9 &   96.7 &    85.6 &   93.5 &                97.4 \\
        Racketsports              &       80.3 &   84.2 &   84.2 &       88.8 &  84.2 &   92.8 &    87.5 &   88.2 &                90.8 \\
        SelfRegulationSCP1        &       77.5 &   78.5 &   76.5 &       85.3 &  90.8 &   69.6 &    93.5 &   82.3 &                78.8 \\
        SelfRegulationSCP2        &       53.9 &   52.2 &   53.3 &       46.1 &  50.6 &   52.8 &    48.3 &   51.7 &                50.6 \\
        StandWalkJump             &       20.0 &   33.3 &   33.3 &       33.3 &  40.0 &   26.7 &    13.3 &   33.3 &                46.7 \\
        UWaveGestureLibrary       &       90.3 &   90.0 &   86.9 &       89.1 &  85.9 &   93.1 &    90.0 &   89.7 &                90.9 \\
        \midrule
        Average Ranks                         &        5.6 &    5.2 &    5.9 &        4.0 &   5.6 &    3.2 &     6.4 &    4.8 &                 4.3 \\
        \bottomrule
    \end{tabular}

	\caption{Results of the signature canonical pipeline along with a selection of classifiers from \citet{ruiz2020benchmarking} (including the top performing MUSE algorithm) with a Random Forest for the UEA archive.}
	\label{tab:all_results_best_rf}
	\end{table}



	Finally, we give in Table \ref{tab:best_rf_hyperparameters} the hyperparameters that were selected for each dataset in the signature pipeline model.

    \begin{table}
	\scriptsize
	\centering
	\begin{tabular}{lccccc}
        \toprule
        {} & \multicolumn{2}{c}{\textbf{Signature hyperparmeters}} & \multicolumn{2}{c}{\textbf{RF hyperparameters}} & \textbf{Other} \\
        \cmidrule(lr){2-3} \cmidrule(lr){4-5} \cmidrule(lr){6-6}
        \textbf{Dataset} & Depth &  Dyadic depth & Max depth &  Num estimators &  Training time (s) \\
        \midrule
        ArticularyWordRecognition &     2 &       2 &            45 &               500 &           60.3 \\
        AtrialFibrillation        &     1 &       2 &          None &                50 &           35.9 \\
        BasicMotions              &     2 &       2 &            24 &               100 &           19.3 \\
        CharacterTrajectories     &     4 &       2 &            80 &               500 &          181.4 \\
        Cricket                   &     2 &       4 &             6 &               500 &          249.0 \\
        DuckDuckGeese             &     1 &       2 &            16 &               100 &          140.9 \\
        ERing                     &     2 &       3 &             8 &              1000 &           16.7 \\
        EigenWorms                &     3 &       3 &            12 &               100 &          250.1 \\
        Epilepsy                  &     2 &       3 &             8 &              1000 &           42.8 \\
        EthanolConcentration      &     2 &       4 &            24 &              1000 &          454.2 \\
        FaceDetection             &     1 &       4 &             8 &              1000 &         1816.2 \\
        FingerMovements           &     1 &       2 &             4 &               100 &           30.8 \\
        HandMovementDirection     &     2 &       2 &          None &                50 &           66.3 \\
        Handwriting               &     6 &       2 &            32 &              1000 &          280.3 \\
        Heartbeat                 &     1 &       4 &          None &                50 &           45.1 \\
        InsectWingbeat            &     1 &       3 &            45 &              1000 &         5367.5 \\
        JapaneseVowels            &     2 &       3 &             6 &              1000 &           95.4 \\
        LSST                      &     4 &       2 &            60 &              1000 &         1590.5 \\
        Libras                    &     6 &       2 &          None &               100 &           28.4 \\
        MotorImagery              &     1 &       3 &            24 &                50 &          347.1 \\
        NATOPS                    &     2 &       3 &            32 &              1000 &           37.8 \\
        PEMS-SF                   &     1 &       3 &            80 &              1000 &          252.3 \\
        PenDigits                 &     3 &       2 &            80 &              1000 &          302.3 \\
        PhonemeSpectra            &     2 &       4 &            45 &              1000 &         2188.7 \\
        RacketSports              &     3 &       2 &          None &               500 &           13.9 \\
        SelfRegulationSCP1        &     3 &       2 &          None &               100 &          186.6 \\
        SelfRegulationSCP2        &     3 &       2 &             6 &                50 &          138.1 \\
        SpokenArabicDigits        &     2 &       3 &            45 &              1000 &         1204.0 \\
        StandWalkJump             &     1 &       3 &             2 &                50 &          101.5 \\
        UWaveGestureLibrary       &     2 &       2 &            60 &               500 &           21.8 \\
        \bottomrule
    \end{tabular}

	\caption{Hyperparameters used for each dataset in the signature pipeline model.}
	\label{tab:best_rf_hyperparameters}
\end{table}






\end{document}